\documentclass[letterpaper]{wato-preprint}

\usepackage{amsmath,amsfonts,bm}

\def\eqref#1{equation~\ref{#1}}

\def\1{\bm{1}}

\DeclareMathAlphabet{\mathsfit}{\encodingdefault}{\sfdefault}{m}{sl}
\SetMathAlphabet{\mathsfit}{bold}{\encodingdefault}{\sfdefault}{bx}{n}

\DeclareMathOperator*{\argmin}{arg\,min}

\usepackage[para,online,flushleft]{threeparttable}
\usepackage{multirow,wrapfig,needspace}
\hypersetup{
  pdftitle={Joint World Action Models through Differentiable Trajectory Optimization},
  pdfauthor={Mintae Kim and Koushil Sreenath}
}

\title{Joint World Action Models through \\Differentiable Trajectory Optimization}
\author{Mintae Kim \& Koushil Sreenath}
\affiliation{\textit{Hybrid Robotics}, BAIR, UC Berkeley}
\correspondence{\textcolor{linkblue}{\{\href{mailto:mintae.kim@berkeley.edu}{mintae.kim}, \href{mailto:koushils@berkeley.edu}{koushils}\}@berkeley.edu}}

\abstract{
Joint world action models (WAMs) generate future actions and states together, modeling world-action trajectories rather than actions alone.
However, trajectory imitation does not ensure that these trajectories are task-effective or consistent with dynamics.
We present \textbf{world action trajectory optimization (WATO)}, which optimizes generated actions and future states jointly for control.
WATO uses both as references in a dynamics-constrained trajectory optimization problem and differentiates the task loss through the optimized trajectory to train the generator.
This makes generated trajectories task-aware and dynamics-consistent while using trajectory optimization only during training.
At deployment, WATO directly generates candidate trajectories and ranks them with a world model, without iterative trajectory optimization.
Across $7$ visual control environments, WATO improves average success from $79.9\%$ for the imitation joint WAM to $91.7\%$, outperforming strong model-based, goal-conditioned RL, and generative control baselines.
Compared with cross-entropy method (CEM), WATO is on average $11.5\times$ faster and uses $600.5\times$ fewer world model evaluations at deployment.
}

\begin{document}
\maketitle

\begin{figure}[H]
\centering
\vspace{.25cm}
\includegraphics[width=\linewidth]{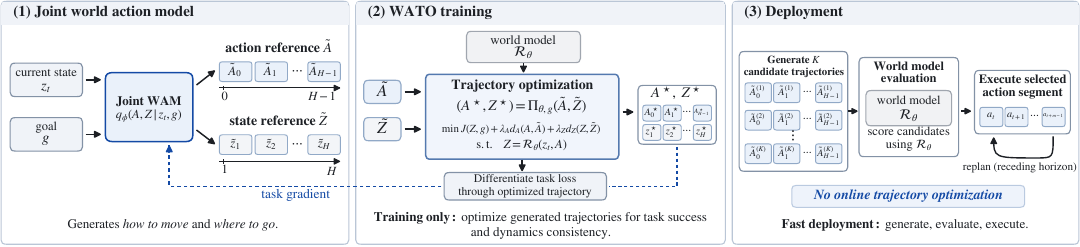}
\caption{
\textbf{WATO overview.}
Joint world-action trajectories are optimized during training and amortized into the generator, enabling deployment with generation and world model evaluation only.
}
\label{fig:wato_concept}
\end{figure}

\section{Introduction}
Generative policies learn distributions over actions or action trajectories from data, reducing online planning \citep{janner2022diffuser,chi2023diffusionpolicy,park2025fql}.
Recent world action models (WAMs) connect action generation with future prediction \citep{zhu2025uwm,ye2026dreamzero}.
We focus on \emph{joint} WAMs that generate future actions and states or observations together.
We treat them as a model class over world-action trajectories, independent of architecture or scale.
Unlike generative policies, joint WAMs specify both how the agent acts and where the trajectory is intended to go.
This provides richer information for control, but does not make the trajectory task-effective.

A joint WAM trained by trajectory imitation mainly reproduces the offline data distribution.
Its generated actions need not reach the goal, and its generated states need not match the consequences of those actions under the control dynamics.
A trajectory can therefore be data-consistent yet task-ineffective or dynamics-inconsistent.
The problem is not only to improve the actions, but to optimize the world-action trajectory as a whole.
This leads to our central question:

\begin{center}
\colorbox{corebg}{
\parbox{0.92\linewidth}{
\vspace{0.35em}
\centering
\textcolor{pubblue}{
\emph{How should joint world-action trajectories be optimized for task performance while remaining consistent with action-conditioned dynamics?}
}
\vspace{0.35em}
}}
\end{center}

Several existing approaches provide natural alternatives, but each uses only part of the world-action trajectory.
Model-based planners optimize actions through world model rollouts~\citep{chua2018deep,kim2026wombet}, but omit generated future states and require online optimization.
Task losses can be backpropagated through learned dynamics~\citep{hafner2020dreamer}, but supervise only the actions.
Consistency objectives can align generated states with world model predictions~\citep{huang2026leflow}, but use them as prediction targets rather than control references.
Optimized trajectories can also be distilled into a policy~\citep{pinneri2021apex}, but detached targets discard how the solution depends on the generated trajectory.
None of these approaches uses both generated actions and future states to parameterize trajectory optimization and learn through its optimized solution.

We introduce \textbf{world action trajectory optimization (WATO)}.
WATO uses generated actions and future states as references in a dynamics-constrained trajectory optimization problem.
The WAM proposes an intended trajectory, while a world model predicts the consequences of its actions \citep{zhou2024dinowm,sobal2025pldm,maes2026lewm,kim2026distribution}.
The task objective drives the trajectory toward the goal.
The action reference specifies how to move, and the state reference specifies where to go.
Trajectory optimization combines both references with the world model to produce a task-directed, dynamics-consistent trajectory.

The generated world-action trajectory does not only initialize the optimizer, but it also parameterizes the optimization objective.
Changing either the actions or future states can therefore change the optimized trajectory.
WATO differentiates the task loss through this solution using implicit differentiation \citep{amos2018diffmpc,wan2024difftori}, providing task gradients to both generated actions and states.
Flow matching provides offline trajectory imitation \citep{lipman2022flowmatching}, while trajectory optimization makes the joint WAM task-aware.

Trajectory optimization is used only during training.
At deployment, WATO directly generates candidate world-action trajectories and evaluates their action sequences with the world model, i.e. action-conditioned dynamics.
Deployment performs a lightweight best-of-$K$ selection over these proposals, but no iterative trajectory optimization is required.

We evaluate WATO on $7$ visual control environments spanning navigation, reaching, contact-rich manipulation, and multi-stage goal-conditioned control.
WATO improves average success from $79.9\%$ for the imitation-trained joint WAM to $91.7\%$, outperforming strong model-based, goal-conditioned RL, and generative control baselines.
Compared with CEM, WATO is $11.5\times$ faster and uses $600.5\times$ fewer world model evaluations.
We also show that differentiating through trajectory optimization outperforms direct rollout, consistency regularization, and distillation.
Removing either the action or state reference reduces performance, showing that both are crucial for control.

Our contributions are threefold:
\begin{itemize}
    \item We study joint WAMs as generative models over world-action trajectories and formulate task-aware world-action generation as dynamics-constrained trajectory optimization.

    \item We introduce \textbf{WATO}, which uses generated actions and future states as optimization references and differentiates the task loss through the optimized trajectory to train the generator.

    \item We show across diverse control tasks that WATO improves joint world-action generation, outperforms strong model-based and learning-based baselines, and amortizes trajectory optimization into the generator for efficient deployment.
\end{itemize}


\section{Preliminaries}
\label{sec:preliminaries}

\textbf{Visual offline goal-conditioned control.}
We consider a reward-free offline dataset of $N$ visual trajectories,
$\mathcal D=\{\tau^{(n)}\}_{n=1}^N$, where
$\tau^{(n)}=(o_0^{(n)},a_0^{(n)},\ldots,o_{T_n}^{(n)})$,
with visual observations $o_t\in\mathcal O$ and primitive actions $a_t\in\mathcal A$.
We group $B$ consecutive actions into an action block
$A_h=(a_{t+hB},\ldots,a_{t+(h+1)B-1})\in\mathcal U$,
where $\mathcal U:=\mathcal A^B$.
An action trajectory $A=(A_0,\ldots,A_{H-1})\in\mathcal U^H$ spans $L=BH$ environment steps. 

We use pretrained latent world models following LeWM~\citep{maes2026lewm}.
The encoder maps an observation to a latent state $z_t=E_\theta(o_t)$, and the action-conditioned predictor rolls out future latent states from an action trajectory.
Given $z_t$ and $A$, we write
\begin{equation}
    \hat Z(A)
    =
    \mathcal R_\theta(z_t,A)
    =
    (\hat z_1,\ldots,\hat z_H).
    \label{eq:latent_rollout}
\end{equation}

A goal observation $o_g$ is encoded as $g=E_\theta(o_g)$.
For a latent trajectory $Z=(z_1,\ldots,z_H)$, we use the terminal goal cost
$J(Z,g)=\|z_H-g\|_2^2$.
Control proceeds in receding horizon by repeatedly generating an action trajectory, executing a segment, and planning from the new observation.

\textbf{Generative policies and joint WAMs.}
Let $c=(z_t,g)$ denote the context that includes the current-state and goal condition.
An action-only generative policy models
$\tilde A\sim q_\phi^A(A\mid c)$.
We focus on \emph{joint} WAMs, which instead model future actions and states together:
\begin{equation}
    (\tilde A,\tilde Z)
    \sim
    q_\phi(A,Z\mid c),
    \qquad
    \tilde Z=(\tilde z_1,\ldots,\tilde z_H).
    \label{eq:world_action_generator}
\end{equation}

A joint WAM therefore induces an action-only distribution through the marginal
$q_{\phi,A}(A\mid c)=\int q_\phi(A,Z\mid c)\,dZ$.
The distinction is the generated object: an action-only policy generates how to act, while a joint WAM additionally generates an intended future trajectory.

The generated future $\tilde Z$ is distinct from the world model prediction
$\hat Z(\tilde A)=\mathcal R_\theta(z_t,\tilde A)$.
The joint WAM proposes both actions and future states, while the pretrained world model predicts the states induced by those actions.
Throughout the paper, $\tilde{\cdot}$ denotes a generated proposal, $\hat{\cdot}$ a world model prediction, and ${\cdot}^\star$ an optimized trajectory.

We parameterize the joint WAM with conditional flow matching~\citep{lipman2022flowmatching}.
For data $x$, noise $\epsilon\sim\mathcal N(0,I)$, condition $c$, and flow time $s\sim\mathrm{Unif}[0,1]$, we use the linear path
$x_s=(1-s)\epsilon+s x$ and train $v_\phi(x_s,s;c)$ toward $x-\epsilon$.
Section~\ref{sec:wato} applies this objective to joint action-state trajectories and makes the resulting generator task-aware through trajectory optimization.


\section{World Action Trajectory Optimization}
\label{sec:wato}

\subsection{Imitating World-Action Trajectories Is Not Enough}
\label{sec:wato_imitation}

We first train a joint world action flow (WAF) to imitate state-action trajectories from the offline dataset.
For a local trajectory, let $X=(A,Z)$ and condition the generator on the current state and goal, $c=(z_t,g)$.
For each training window, we use its final observation as the goal.
The WAF generates actions and future states jointly as
$(\tilde A,\tilde Z)\sim q_\phi(A,Z\mid z_t,g)$.

We parameterize $q_\phi$ with conditional flow matching.
Let $\epsilon=(\epsilon_A,\epsilon_Z)\sim\mathcal N(0,I)$,
$s\sim\mathrm{Unif}[0,1]$, and $X_s=(1-s)\epsilon+sX$.
The joint vector field $v_\phi=(v_\phi^A,v_\phi^Z)$ is trained with
\begin{equation}
    \mathcal L_{\mathrm{WAF}}
    =
    \mathbb E
    \left[
        \operatorname{MSE}
        \left(
            v_\phi^A(X_s,s;c),
            A-\epsilon_A
        \right)
        +
        \lambda_{\mathrm{WAF}}
        \operatorname{MSE}
        \left(
            v_\phi^Z(X_s,s;c),
            Z-\epsilon_Z
        \right)
    \right].
    \label{eq:wato_fm}
\end{equation}

This objective imitates the joint state-action distribution in the offline data, but does not optimize trajectories for control.
The generated actions need not reach the goal under the pretrained dynamics, so
$J(\mathcal R_\theta(z_t,\tilde A),g)$ can remain large.
The generated states also need not match the consequences of those actions, so
$\tilde Z$ need not equal $\mathcal R_\theta(z_t,\tilde A)$.
Imitation alone therefore provides neither task optimality nor per-sample dynamics-consistency.

\subsection{Why Not Optimize the Generator Directly?}
\label{sec:wato_alternatives}

A simple way to provide task-awareness is to evaluate its generated actions through the world model.
Direct rollout training minimizes
$\mathcal L_{\mathrm{rollout}}
=
J(\mathcal R_\theta(z_t,\tilde A),g)$.
This provides a task gradient to $\tilde A$, but $\tilde Z$ does not affect action optimization.
The joint WAF is thus used only as an action generator.

A stronger variant can also match the generated states to the world model rollout.
For example, one can minimize
$\mathcal L_{\mathrm{direct+cons}}
=
J(\mathcal R_\theta(z_t,\tilde A),g)
+
\beta d_Z(\mathcal R_\theta(z_t,\tilde A),\tilde Z)$.
This encourages dynamics-consistency, but still trains $\tilde Z$ to follow the rollout rather than to guide trajectory optimization.

Another alternative is optimization distillation.
Given a generated proposal $\tilde\tau=(\tilde A,\tilde Z)$, a trajectory optimizer produces an optimized trajectory $\tau^\star$, which is then treated as a detached regression target.
This transfers optimized trajectories to the generator, but discards how $\tau^\star$ changes with $\tilde\tau$.
The optimizer provides targets rather than gradients through the optimized solution.

These alternatives either leave the generated future outside trajectory optimization or use optimization only to produce detached targets.
WATO instead uses the full generated state-action trajectory to define the optimization problem and learns through its optimized solution.

\subsection{Dynamics-Constrained Trajectory Projection}
\label{sec:wato_projection}

WATO uses the generated trajectory as a reference for trajectory optimization rather than executing it directly.
For the current state $z_t$, define the set of trajectories consistent with the pretrained world model as
$\mathcal F_\theta(z_t)
=
\{(A,Z):A\in\mathcal U^H,\,
Z=\mathcal R_\theta(z_t,A)\}$.
Given a generated reference $\tilde\tau=(\tilde A,\tilde Z)$, we define the \textcolor{pubblue}{\emph{dynamics-constrained trajectory projection}}
\begin{equation}
\colorbox{corebg}{
\centering
$\displaystyle
\Pi_{\theta,g}(\tilde\tau)
=
\operatorname*{arg\,min}_{(A,Z)\in\mathcal F_\theta(z_t)}
J(Z,g)
+
\lambda_A d_A(A,\tilde A)
+
\lambda_Z d_Z(Z,\tilde Z)
$
}
\label{eq:wato_projection}
\end{equation}
The task cost drives the trajectory toward the goal.
The action reference preserves how the WAM proposes to move, while the state reference preserves where it proposes to go.
The world model constrains these references through action-conditioned dynamics.

We use
$d_A(A,\tilde A)
=
H^{-1}\sum_{h=0}^{H-1}\|A_h-\tilde A_h\|_2^2$
and
$d_Z(Z,\tilde Z)
=
H^{-1}\sum_{h=1}^{H}
\|\mathcal N_Z(z_h)-\mathcal N_Z(\tilde z_h)\|_2^2$,
where $\mathcal N_Z$ is a fixed latent standardization.
We solve~\Cref{eq:wato_projection} in reduced space, with the action trajectory as the only optimization variable and
$Z(A)=\mathcal R_\theta(z_t,A)$.
The optimization becomes
\begin{equation}
    A^\star
    =
    \operatorname*{arg\,min}_{A\in\mathcal U^H}
    J\!\left(Z(A),g\right)
    +
    \lambda_A d_A(A,\tilde A)
    +
    \lambda_Z d_Z(Z(A),\tilde Z),
    \qquad
    Z^\star
    =
    Z(A^\star).
    \label{eq:wato_trajectory_optimization}
\end{equation}

The optimized trajectory is task-directed and dynamics-consistent while remaining close to the generated state-action trajectory.
Importantly, $\tilde Z$ is a reference, not an independently optimized state trajectory.
Dynamics-consistency follows from the world model rollout rather than an additional penalty.
Because both $\tilde A$ and $\tilde Z$ parameterize the optimization objective, changing either reference can change the optimized trajectory.
We next characterize this dependence and show how generated future states shape action optimization through the learned dynamics.

\subsection{Learning Through the Optimized Trajectory}
\label{sec:wato_diff}

The generated trajectory parameterizes the trajectory optimization problem, not only its initialization.
Define the task objective
$F(A)=J(\mathcal R_\theta(z_t,A),g)$
and the standardized rollout
$r(A)=\operatorname{vec}\mathcal N_Z(\mathcal R_\theta(z_t,A))$.
We similarly write
$\tilde r=\operatorname{vec}\mathcal N_Z(\tilde Z)$.
The reduced objective in~\Cref{eq:wato_trajectory_optimization} is
$\Psi(A;\tilde A,\tilde r)
=
F(A)
+
\lambda_A H^{-1}\|A-\tilde A\|_2^2
+
\lambda_Z H^{-1}\|r(A)-\tilde r\|_2^2$.

Let
$S_\star=\partial r(A)/\partial A|_{A=A^\star}$
denote the standardized world model rollout Jacobian.
For an interior local solution, the first-order condition is
\begin{equation}
    \nabla F(A^\star)
    +
    \frac{2\lambda_A}{H}
    (A^\star-\tilde A)
    +
    \frac{2\lambda_Z}{H}
    S_\star^\top
    \bigl(r(A^\star)-\tilde r\bigr)
    =
    0.
    \label{eq:wato_optimality}
\end{equation}
The action reference acts directly in action space.
The state reference acts through $S_\star^\top$, which maps a future-state error into an action-space optimization direction.
Generated future states therefore affect trajectory optimization through the action-conditioned world dynamics.

\noindent
\textcolor{pubblue}{\textbf{Proposition 1 (Dynamics-aware reference sensitivity).}}
Suppose $A^\star$ is an interior local solution, $\Psi$ is twice continuously differentiable near $A^\star$, and
$\mathcal H_\star
:=
\nabla_{AA}^2
\Psi(A^\star;\tilde A,\tilde r)$
is nonsingular.
Then
\begin{equation}
\colorbox{corebg}{
$\displaystyle
    \frac{\partial A^\star}{\partial\tilde A}
    =
    \frac{2\lambda_A}{H}
    \mathcal H_\star^{-1},
    \qquad
    \frac{\partial A^\star}{\partial\tilde r}
    =
    \frac{2\lambda_Z}{H}
    \mathcal H_\star^{-1}S_\star^\top
$}
\label{eq:wato_reference_sensitivity}
\end{equation}

Actions affect the optimized trajectory through the local optimization Hessian.
Future states additionally pass through the rollout Jacobian, so their influence depends on how actions change the predicted future.
Both references can thus receive task gradients through the optimized solution.

This structure also changes the local geometry of action optimization.
The state-reference term contributes
$2S_\star^\top S_\star$
plus a residual-weighted second-order term from the rollout.
When the state-reference residual is small, the local Hessian is approximated by
\begin{equation}
    \mathcal H_\star
    \approx
    \nabla^2 F(A^\star)
    +
    \frac{2\lambda_A}{H}I
    +
    \frac{2\lambda_Z}{H}S_\star^\top S_\star.
    \label{eq:wato_local_geometry}
\end{equation}

\noindent
\textcolor{pubblue}{\textbf{Corollary 1 (Dynamics-induced local geometry).}}
The action reference contributes isotropic curvature in action space, while the state reference contributes the positive-semidefinite term $S_\star^\top S_\star$.
Action perturbations are therefore weighted by how strongly they change the predicted future.
Generated states thus shape the local geometry of trajectory optimization through the learned dynamics.

We compute these sensitivities by implicit differentiation of the trajectory optimality conditions.
Further derivations and local analysis are given in Appendix~\ref{app:theory}.

\subsection{The WATO Objective}
\label{sec:wato_objective}

Let
$\tilde\tau_\phi=(\tilde A,\tilde Z)\sim q_\phi(A,Z\mid z_t,g)$
denote a generated trajectory and
$\tau^\star=(A^\star,Z^\star)=\Pi_{\theta,g}(\tilde\tau_\phi)$
its optimized trajectory.
WATO trains the same joint WAF with both imitation and task supervision:
\begin{equation}
\colorbox{corebg}{
$\displaystyle
\mathcal L_{\mathrm{WATO}}
=
\mathcal L_{\mathrm{WAF}}
+
\lambda_{\mathrm{task}}
\mathbb E
\left[
    J(Z^\star,g)
\right]
$}
\label{eq:wato_loss}
\end{equation}
The flow matching term continues to imitate offline state-action trajectories, while the task term favors generated references that lead to better optimized trajectories.

The task gradient passes through the optimized trajectory as
\begin{equation}
    \frac{d J(Z^\star,g)}{d\phi}
    =
    \frac{\partial J}{\partial\tau^\star}
    \underbrace{
        \textcolor{pubblue}{
        \frac{\partial\tau^\star}{\partial\tilde\tau_\phi}
        }
    }_{\mathclap{\text{\scriptsize solution sensitivity}}}
    \frac{\partial\tilde\tau_\phi}{\partial\phi}.
    \label{eq:wato_task_gradient}
\end{equation}
By Proposition~1, action gradients pass through the local optimization Hessian, while state gradients additionally pass through the rollout Jacobian.
The world model is differentiable when computing these gradients, but its parameters $\theta$ remain fixed.
WATO updates only the generator parameters $\phi$.

\subsection{Training and Deployment}
\label{sec:wato_training}

WATO uses two training stages.
We first pretrain the joint WAF with $\mathcal L_{\mathrm{WAF}}$ to imitate offline state-action trajectories.
We then continue training the same generator with $\mathcal L_{\mathrm{WATO}}$, differentiating through trajectory optimization to provide task supervision.

Trajectory optimization is removed after training.
At deployment, WATO generates $K$ independent proposals
$(\tilde A^{(k)},\tilde Z^{(k)})\sim q_\phi(A,Z\mid z_t,g)$.
Each action proposal is rolled out with the pretrained world model as
$\hat Z^{(k)}=\mathcal R_\theta(z_t,\tilde A^{(k)})$.
The controller selects
$k^\star=\operatorname*{arg\,min}_k J(\hat Z^{(k)},g)$
and executes the selected action segment in receding horizon.
The generated state trajectory is not directly scored at deployment.
It contributes through joint generation and trajectory-optimization training.

Optimization is used to train the generator, \emph{not} to operate the deployed controller.
Deployment requires only trajectory generation and world model evaluation.
Full algorithm is provided in \Cref{alg:wato}.

\paragraph{\textcolor{pubblue}{Remark: Optimization amortization.}}
Under local regularity assumptions, small projection residuals bound the task-cost gap between generated and optimized trajectories and the generated trajectory's inconsistency with the world model (see Appendix~\ref{app:optimization_amortization}).

\begin{algorithm}[t!]
\caption{\textcolor{pubblue}{World Action Trajectory Optimization (WATO)}}
\label{alg:wato}
\begin{algorithmic}
\footnotesize

\Require Offline dataset $\mathcal D$, pretrained world model $\mathcal R_\theta$ (\Cref{eq:latent_rollout})

\LComment{\color{pubblue} WAF pretraining}
\State Pretrain joint WAF $q_\phi(A,Z\mid z_t,g)$ with flow matching
(\Cref{eq:wato_fm})

\LComment{\color{pubblue} WATO training loop}
\While{not converged}
    \State Sample $(z_t,g,A,Z)\sim\mathcal D$
    \State Generate $(\tilde A,\tilde Z)\sim q_\phi(A,Z\mid z_t,g)$ (\Cref{eq:world_action_generator})

    \BeginBox[fill=white]
    \LComment{Dynamics-constrained trajectory optimization}
    \State Compute
    $(A^\star,Z^\star)
    =
    \Pi_{\theta,g}(\tilde A,\tilde Z)$
    (\Cref{eq:wato_projection})
    \EndBox
    \BeginBox[fill=white]
    \LComment{Learning through the optimized trajectory}
    \State Compute task loss $J(Z^\star,g)$
    \State Differentiate through the optimized trajectory
    (\Cref{eq:wato_task_gradient})
    \State Update $\phi$ with $\mathcal L_{\mathrm{WATO}}$
    (\Cref{eq:wato_loss})
    \EndBox
\EndWhile

\LComment{\color{pubblue} Deployment}
\State Generate $K$ proposals
$(\tilde A^{(k)},\tilde Z^{(k)})
\sim q_\phi(A,Z\mid z_t,g)$
\State Roll out
$\hat Z^{(k)}
=
\mathcal R_\theta(z_t,\tilde A^{(k)})$
for each proposal
\State Select
$k^\star
=
\argmin_k J(\hat Z^{(k)},g)$
\State Execute the selected action segment and replan

\end{algorithmic}
\end{algorithm}


\section{Related Work}
\label{sec:related_work}

\paragraph{Generative policies and WAMs.}
Generative policies learn distributions over actions or trajectories with multimodality \citep{janner2022diffuser,ajay2022decisiondiffuser,chi2023diffusionpolicy,park2025fql}.
WAMs extend action generation by jointly modeling future actions and states or observations \citep{zhu2025uwm,cen2025worldvla,ye2026dreamzero}.
Recent work also studies how generated futures should be used at inference time \citep{yuan2026fastwam,zhao2026fasterwam}.
WATO instead optimizes the generated world-action trajectory for control.
Generated actions and future states jointly define a dynamics-constrained optimization problem, and the generator is trained through its solution.

\paragraph{World models and model-based planning.}
World models learn action-conditioned dynamics and use predicted futures to evaluate and/or optimize actions \citep{hafner2019dream,hansen2024td,sobal2025pldm,maes2026lewm}.
Model-based planners commonly optimize action sequences online using sampling or gradient-based methods \citep{power2022flowmppi,zhou2024dmpc}.
WATO uses a pretrained world model to constrain trajectory optimization during training.
Deployment only requires trajectory generation and world model evaluation, without online optimization.

\paragraph{Differentiable optimization and generative control.}
Differentiable optimization exposes how an optimized trajectory changes with model, cost, or policy parameters \citep{amos2018diffmpc,amos2019dcem}.
Prior work uses this principle to train policies through optimization or predictive control \citep{xu2024leto,xu2025diffog,wan2024difftori}.
Other methods combine generative models with trajectory optimization, model-based projections, or MPC
\citep{romer2024diffusion,mizuta2025cfmmppi,idoko2025flowopt,li2026hardflow}.
WATO uses both actions and future states to define the optimization objective, then differentiates task performance through its solution.
Thus, differentiable optimization serves as a learning mechanism for WAMs rather than refinement only.

\paragraph{Offline goal-conditioned control.}
Offline goal-conditioned RL (GCRL) learns to reach goals from fixed trajectory data.
Prior methods include goal-conditioned behavior cloning, value learning, quasimetric learning, and hierarchical RL \citep{ding2019gcbc,kostrikov2021gcivl,wang2023qrl,park2023hiql,park2025ogbench}.
WATO addresses the same downstream goal-reaching problem from a different direction.
It starts from a joint WAF and makes its trajectories task-aware using a world model and differentiable trajectory optimization.
We compare against these methods as goal-conditioned control baselines.


\section{Experiments}
\label{sec:experiments}

We evaluate WATO along four questions.
First, does world-action trajectory optimization improve goal-conditioned control over model-based, generative, and RL methods?
Second, is differentiating through trajectory optimization more effective than direct task training or optimization distillation?
Third, do generated future states improve control beyond action generation alone?
Finally, is trajectory optimization successfully transferred into the generator during training?

\subsection{Experimental Setup}
\label{sec:experimental_setup}

\paragraph{Environments.}
We evaluate WATO on $7$ environments covering distinct control challenges.
\texttt{TwoRooms} tests visual navigation, \texttt{Reacher} tests continuous reaching, and \texttt{PushT} tests contact-rich pushing \citep{maes2026lewm}.
We additionally use four visual manipulation tasks from OGBench: \texttt{cube-\{single, double\}}, \texttt{scene-play}, and \texttt{puzzle-3x3} \citep{park2025ogbench}.
Together, these environments cover navigation, reaching, manipulation, and multi-stage goal-conditioned control.

\paragraph{Evaluation.}
We evaluate goal reaching from offline data.
Each initial observation is paired with a goal observation $L=25$ primitive steps later in a evaluation trajectory, and the controller receives a $50$-step execution budget.
Unless otherwise stated, WATO generates $K=8$ candidates using $8$ flow integration steps (NFE $=8$) at deployment.
Success rate is the primary metric.
We also report final goal error, latency, and world model evaluations in the appendix.

\paragraph{Baselines.}
We compare against three groups of methods.
For \emph{goal-conditioned learning}, we use GCBC~\citep{ding2019gcbc}, GCIVL~\citep{kostrikov2021gcivl}, GCIQL~\citep{park2025ogbench}, and QRL~\citep{wang2023qrl}.
For \emph{generative control}, we compare with an flow policy, a joint WAF trained only by flow matching, and generative predictive control (GPC)~\citep{kurtz2025gpc}, adapted to our offline setting.
For \emph{model-based control}, we use CEM~\citep{maes2026lewm} and gradient-based MPC.
GPC, CEM, gradient-based MPC, and WATO use the same pretrained LeWM.
Model-based planners also use the same task cost.
All methods use the same data, goal distribution, and evaluation protocol.
Training details are given in Appendix~\ref{app:baseline_training}.

\subsection{How Good is WATO for Visual Offline Goal-Conditioned Control?}
\label{sec:main_results}

We first evaluate WATO across all $7$ environments.
Table~\ref{tab:main_results} reports the success rates.
WATO achieves the highest macro-average success of $91.7\%$ and is best or tied for best on six of $7$ environments, while GPC performs best on \texttt{Reacher}.

Compared with the joint WAF, WATO improves average success from $79.9\%$ to $91.7\%$.
This $11.8$ percentage points (pp) gain shows that trajectory imitation alone does not fully exploit the generated state-action trajectory.
Optimizing it during training makes the generator more effective for control.

WATO also outperforms online model-based planning.
CEM and gradient-based MPC optimize actions at deployment using the same world model, achieving average success rates of $65.0\%$ and $76.7\%$, respectively.
WATO uses optimization only during training and deploys the generator without online optimization.
Its gains therefore come from improving the generator rather than additional extensive test-time search.
Section~\ref{sec:amortization_results} examines this transfer of optimization into the generator.

\newcommand{\std}[1]{{\tiny $\pm #1$}}

\begin{table}[!htbp]
\centering
\caption{
\textbf{Overall control performance.}
Success rates across $\mathbf{7}$ visual control environments.
}
\label{tab:main_results}
\scriptsize
\setlength{\tabcolsep}{5.0pt}
\begin{threeparttable}
\begin{tabular}{lcccccccccc}
\toprule
& \multicolumn{4}{c}{Goal-conditioned learning}
& \multicolumn{3}{c}{Generative control}
& \multicolumn{2}{c}{Model-based control}
& \\
\cmidrule(lr){2-5}
\cmidrule(lr){6-8}
\cmidrule(lr){9-10}
Environment
& GCBC
& GCIVL
& GCIQL
& QRL
& Flow Policy
& Joint WAF
& GPC
& CEM
& Grad. MPC
& \textcolor{pubblue}{\textbf{WATO}} \\
\midrule
\texttt{TwoRooms}      & $81$ \std{2} & $77$ \std{3} & $80$ \std{5} & $85$ \std{2} & $79$ \std{2} & $83$ \std{1} & $87$ \std{2} & $86$ \std{3} & $\mathbf{92}$ \std{3} & $\mathbf{93}$ \std{0} \\
\texttt{Reacher}       & $74$ \std{3} & $65$ \std{3} & $71$ \std{4} & $73$ \std{3} & $64$ \std{5} & $67$ \std{3} & $\mathbf{91}$ \std{2} & $70$ \std{2} & $81$ \std{4} & $85$ \std{2} \\
\texttt{PushT}         & $66$ \std{1} & $45$ \std{2} & $39$ \std{0} & $58$ \std{1} & $77$ \std{2} & $84$ \std{4} & $\mathbf{99}$ \std{1} & $82$ \std{2} & $97$ \std{3} & $\mathbf{100}$ \std{0} \\
\texttt{cube-single}   & $75$ \std{0} & $45$ \std{3} & $52$ \std{1} & $72$ \std{3} & $81$ \std{3} & $88$ \std{1} & $78$ \std{3} & $68$ \std{8} & $76$ \std{2} & $\mathbf{99}$ \std{1} \\
\texttt{cube-double}   & $63$ \std{1} & $21$ \std{4} & $37$ \std{5} & $60$ \std{1} & $62$ \std{4} & $\mathbf{89}$ \std{2} & $64$ \std{7} & $58$ \std{3} & $71$ \std{1} & $\mathbf{90}$ \std{3} \\
\texttt{scene-play}    & $55$ \std{2} & $32$ \std{2} & $40$ \std{4} & $58$ \std{0} & $68$ \std{3} & $78$ \std{3} & $37$ \std{2} & $42$ \std{11} & $59$ \std{5} & $\mathbf{94}$ \std{1} \\
\texttt{puzzle-3x3}    & $51$ \std{2} & $49$ \std{8} & $37$ \std{6} & $45$ \std{1} & $54$ \std{3} & $70$ \std{3} & $39$ \std{5} & $49$ \std{4} & $61$ \std{2} & $\mathbf{81}$ \std{2} \\
\midrule
Average
& $66.4$
& $47.7$
& $50.9$
& $64.4$
& $69.3$
& $79.9$
& $70.7$
& $65.0$
& $76.7$
& $\mathbf{91.7}$ \\
\bottomrule
\end{tabular}
\begin{tablenotes}
\item[1] Environments except for the top three are from OGBench~\citep{park2025ogbench} \texttt{visual-v0}.
\item[2] Success rates are mean $\pm$ std (\%) over $\mathbf{4}$ seeds, with $\mathbf{100}$ tests per seed. Bold indicates the best mean and results within $2$ pp of it.
\end{tablenotes}
\end{threeparttable}
\end{table}

\subsection{What Does WATO Generate?}
\label{sec:qualitative_control}

\begin{figure}[t]
    \centering
    \vspace{0.5cm}
    \includegraphics[width=\linewidth]{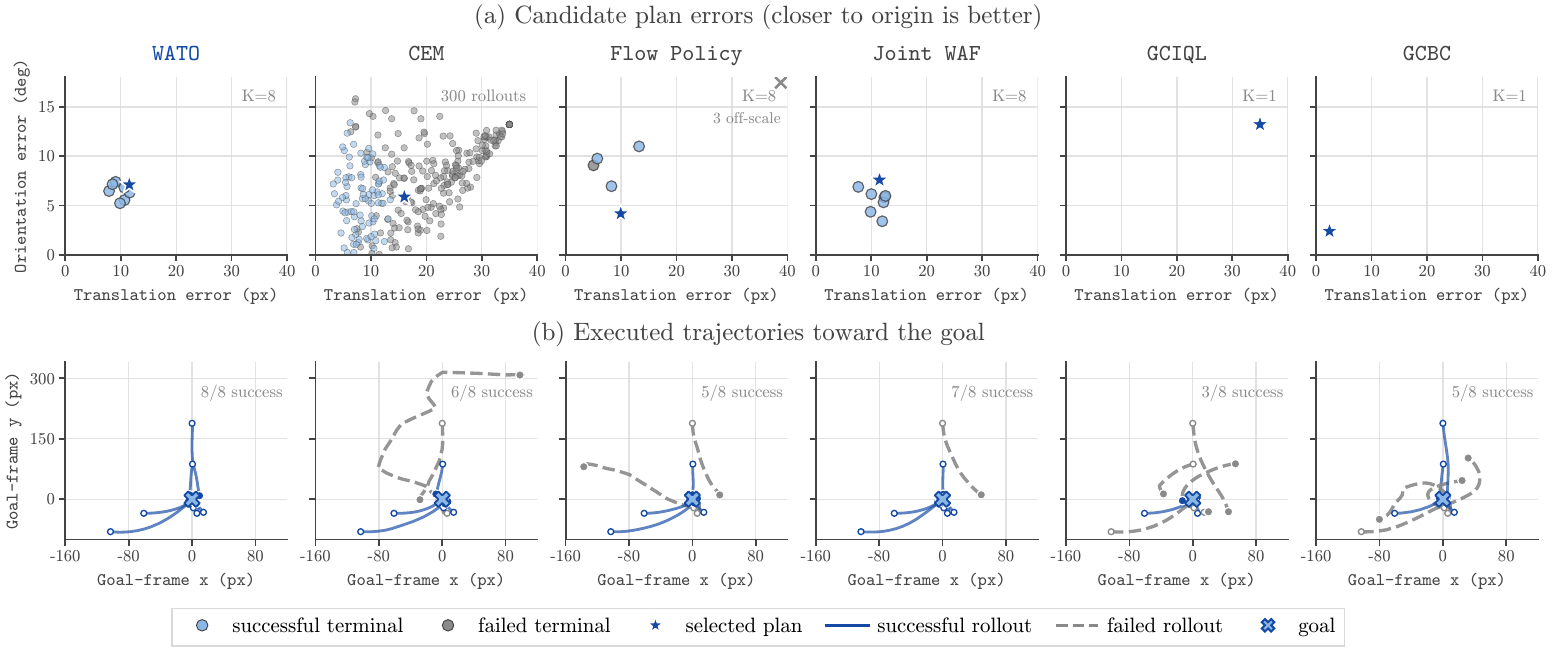}
    \caption{
    \textbf{Visualizing control behavior in \texttt{PushT}.}
    \emph{(a)} Terminal object-position and orientation errors for candidates from a \emph{single planning step} in one representative episode.
    Each candidate is replayed from the same state.
    \emph{(b)} Goal-aligned object trajectories over $8$ \emph{full episodes} with heterogeneous initializations, with the goal at the origin.
    }
    \label{fig:pusht_mechanism}
\end{figure}

Figure~\ref{fig:pusht_mechanism} separates single-step candidate quality from full closed-loop behavior.
A good local proposal does not necessarily yield reliable control over repeated planning.

At one planning step, WATO generates only $K=8$ concentrated, low-error candidates.
CEM uses $300$ rollouts but produces a broader candidate set, while the flow policy includes several large-error proposals.
GCIQL and GCBC provide only a single proposal without candidate selection.
WATO therefore achieves high-quality candidates with little test-time search.

Across eight heterogeneous episodes, WATO succeeds in $8/8$, compared with $7/8$ for the joint WAF, $6/8$ for CEM, and $5/8$ for the flow policy.
Relative to the joint WAF, WATO yields a tighter candidate distribution and more reliable closed-loop trajectories.
CEM also shows that more online search does not necessarily improve reliability.
GCBC further illustrates the gap between local proposal quality and closed-loop behavior: its proposal in Fig.~\ref{fig:pusht_mechanism}(a) can lie near the goal, yet only $5/8$ episodes succeed in Fig.~\ref{fig:pusht_mechanism}(b).
Overall, WATO combines sample-efficient candidate generation with stable repeated planning.

\Needspace{15\baselineskip}
\subsection{How Beneficial is Differentiation Through Trajectory Optimization?}
\label{sec:wato_training_ablation}

\begin{wrapfigure}{r}{0.42\linewidth}
    \centering
    \includegraphics[width=0.95\linewidth]{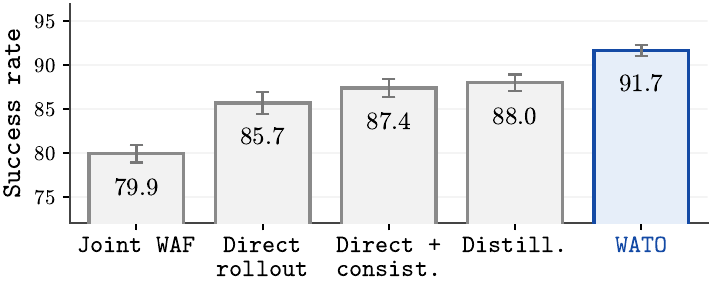}
    \caption{
    \textbf{Training objective ablation.}
    Average success across all environments.
    }
    \label{fig:wato_training_ablation}

\end{wrapfigure}

We next ask whether WATO benefits specifically from differentiating through trajectory optimization.
All variants use the same joint WAF, offline data, and world model.
Only the training objective changes.
\emph{Joint WAF} uses flow matching alone.
\emph{Direct rollout} adds the goal-reaching loss through the world model rollout, supervising the generated actions.
\emph{Direct rollout + consistency} also matches states to the rollout.
\emph{Distillation} trains the generator toward detached trajectories produced by the same optimizer.
WATO instead differentiates through the optimized solution, allowing both action and state references to receive gradients according to how they affect that solution.

Figure~\ref{fig:wato_training_ablation} shows a steady improvement across these objectives.
Joint WAF achieves $79.9\%$, direct rollout $85.7\%$, direct rollout with consistency $87.4\%$, and distillation $88.0\%$.
WATO reaches $91.7\%$, indicating that differentiating through optimization provides more effective supervision than direct task losses, consistency regularization, or detached optimized targets.

\subsection{Is Trajectory Optimization Well-Amortized through WATO?}
\label{sec:amortization_results}

\noindent
\begin{minipage}[t]{0.52\linewidth}
\vspace{0pt}

WATO uses trajectory optimization during training but removes it at deployment.
We test whether this optimization is transferred into the generator by applying the same trajectory optimization layer before and after WATO training.

Table~\ref{tab:wato_amortization} reports the additional success obtained from online trajectory optimization.
Before WATO training, optimization improves success by $10.4$ pp on average.
After WATO training, the remaining gain drops to $1.3$ pp.
Thus, WATO learns to generate goal-reaching and dynamics-consistent trajectories that already capture most of the benefit of trajectory optimization.

\end{minipage}
\hfill
\begin{minipage}[t]{0.45\linewidth}
\captionsetup{type=table}
\vspace{0.15em}
\centering
\scriptsize
\setlength{\tabcolsep}{4.5pt}
\begin{tabular}{lcc}
\toprule
Environment
& Before WATO
& After WATO \\
\midrule
\texttt{TwoRooms}      & $+8.0$  & $+1.0$ \\
\texttt{Reacher}       & $+15.0$ & $+3.0$ \\
\texttt{PushT}         & $+14.0$ & $+0.0$ \\
\texttt{cube-single}   & $+9.0$  & $+1.0$ \\
\texttt{cube-double}   & $+3.0$  & $+1.0$ \\
\texttt{scene-play}    & $+14.0$ & $+1.0$ \\
\texttt{puzzle-3x3}    & $+10.0$ & $+2.0$ \\
\midrule
Average                & $\mathbf{+10.4}$ & $\mathbf{+1.3}$ \\
\bottomrule
\end{tabular}
\caption{
\textbf{Remaining benefit of online trajectory optimization is small.}
Additional success from the same optimizer before and after WATO training, in percentage points.
}
\label{tab:wato_amortization}
\end{minipage}

\subsection{Should We Generate Both States and Actions?}
\label{sec:wato_reference_ablation}

We next isolate the roles of generated actions and future states.
The \emph{action-only} variant removes $d_Z$, while the \emph{state-only} variant removes $d_A$.
They achieve average success rates of $85.7\%$ and $85.0\%$, respectively, compared with $91.7\%$ for full WATO.
Thus, neither reference alone recovers the performance of their combination.
Generated future states therefore contribute directly to trajectory optimization rather than serving only as auxiliary predictions.
This result is consistent with Proposition~1: state references influence optimized actions through the world model rollout Jacobian.
We further compare generated and rollout-derived state references to test whether generated futures provide benefits beyond dynamics-aware regularization (Appendix~\ref{app:full_wato_ablations}).

\subsection{How Does Trajectory Optimization Shape the Generator?}
\label{sec:wato_correction}

\begin{figure}[!htbp]
    \centering
    \includegraphics[
        width=.75\linewidth
    ]{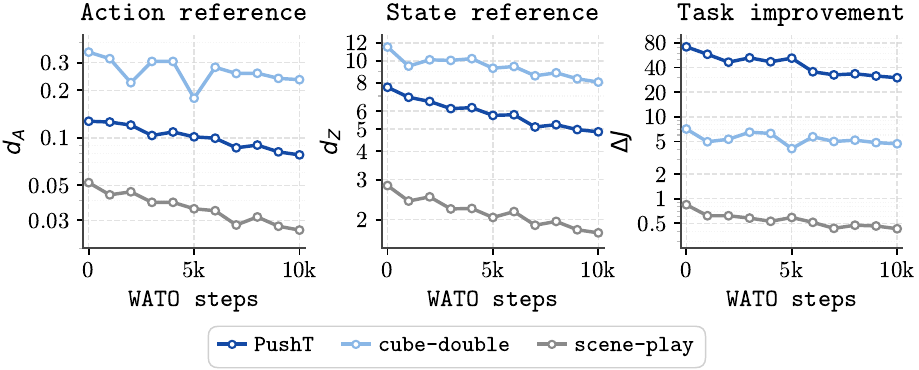}
    \caption{
    \textbf{Trajectory projection during WATO training.}
    Training dynamics on three representative environments, evaluated every $1{,}000$ WATO steps.
    }
    \label{fig:wato_correction}
\end{figure}

We next examine how the generator changes during WATO training on three environments: \texttt{PushT}, \texttt{cube-double}, and \texttt{scene-play}.
For each generated world-action trajectory, we measure
$d_A(A^\star,\tilde A)$,
$d_Z(Z^\star,\tilde Z)$,
and the task improvement
$\Delta J = J(\mathcal R_\theta(z_t,\tilde A),g)-J(Z^\star,g)$
(see Sec.~\ref{sec:wato_projection} for definitions).

Figure~\ref{fig:wato_correction} shows that all three metrics decrease despite some fluctuations.
From initialization to $10$k WATO steps, the action-reference distance decreases by $39.3\%$ on average across the three environments, the state-reference distance by $30.5\%$, and the remaining goal-reaching task improvement by $31.8\%$.
The generator therefore moves closer to its optimized trajectories while requiring progressively less correction from trajectory optimization.
This provides direct evidence that trajectory optimization is amortized into world-action generation.

\Needspace{17\baselineskip}
\subsection{How Efficient is WATO at Deployment?}
\label{sec:wato_efficiency}

\begin{wraptable}[13]{r}{0.37\linewidth}
    \centering
    \scriptsize
    \setlength{\tabcolsep}{5pt}
    \begin{tabular}{lcc}
    \toprule
    Environment
    & Latency
    & WM evals \\
    \midrule
    TwoRooms       & $18.9\times$ & $1125\times$ \\
    Reacher        & $13.6\times$ & $1125\times$ \\
    PushT          & $25.0\times$ & $1125\times$ \\
    cube-single    & $5.8\times$  & $375\times$ \\
    cube-double    & $6.3\times$  & $375\times$ \\
    scene-play     & $12.2\times$ & $375\times$ \\
    puzzle-3x3     & $9.2\times$  & $375\times$ \\
    \midrule
    Geometric mean     & $\mathbf{11.5\times}$ & $\mathbf{600.5\times}$ \\
    \bottomrule
    \end{tabular}
    \caption{
    \textbf{Efficiency relative to CEM at deployment.}
    Ratios are $T_{\mathrm{CEM}}/T_{\mathrm{WATO}}$ and
    $N_{\mathrm{CEM}}/N_{\mathrm{WATO}}$. Higher is better.
    }
    \label{tab:wato_efficiency}
\end{wraptable}

Table~\ref{tab:wato_efficiency} compares WATO with CEM at deployment.
WATO reduces latency in every environment, with speedups from $5.8\times$ to $25.0\times$ and a geometric mean of $11.5\times$.
World model evaluations decrease more sharply, from $375\times$ to $1125\times$, with a geometric mean of $600.5\times$.

The larger reduction in world model evaluations than in latency shows that model evaluations alone do not determine wall-clock time, because trajectory generation and execution remain.
WATO and the joint WAF use the same deployment procedure, so WATO's control gains require no additional test-time computation.
Absolute latency, evaluation counts, and timing details are reported in Appendix~\ref{app:deployment_efficiency}.


\section{Discussion and Conclusion}
\label{sec:discussion}

WATO shows that trajectory optimization can train joint WAMs rather than only refine actions at deployment.
Across $7$ environments, it improves average success from $79.9\%$ for the joint WAF to $91.7\%$, with ablations showing that both action and state references contribute.
Decreasing projection residuals and optimization benefit further indicate that WATO transfers trajectory optimization into the generator.
Generated future states also play a direct control role: as trajectory references, they shape optimized actions through the world model dynamics.

World models and joint WAMs therefore play complementary roles.
The joint WAM proposes trajectories, while the pretrained world model provides action-conditioned dynamics during training and evaluates actions at deployment.
WATO uses this model to improve generation without iterative online planning.
Limitations include dependence on learned dynamics, the finite-horizon goal-conditioned setting, and evaluation only with flow matching.
Extending WATO to robust longer-horizon control and other joint generative models remains open.



\section*{AI Use Statement}
Generative AI was \emph{only} used to assist with grammar checking and writing polish.
\emph{All ideas, scientific content, experimental results, and conclusions are the authors' own.}

\section*{Reproducibility Statement}
Implementation details, model hyperparameters, optimization procedures, and evaluation protocols are provided in the appendix.
We also include anonymous code to reproduce the experiments.


\bibliographystyle{plainnat}
\bibliography{reference}


\clearpage
\beginappendix

\section{World Action Models as a Generative Control Model Class}
\label{app:wam_control_class}

We make this \textcolor{pubblue}{\emph{perspective manifest}} by treating WAMs as a mathematical class of conditional trajectory models, independent of architecture or scale.
Let $c=(z_t,g)$ denote the current-state and goal condition.
An action-only generative policy models $q_\phi^A(A\mid c)$, a world model represents action-conditioned dynamics $p_\theta(Z\mid A,c)$, and a joint WAM models $q_\phi(A,Z\mid c)$.
What differs is the modeled object.
A joint WAM generates both an action trajectory and an intended future trajectory, with action marginal
$q_{\phi,A}(A\mid c)=\int q_\phi(A,Z\mid c)\,dZ$.

\paragraph{Model class and scale are separate choices.}
A joint WAM does not require a particular architecture, generative parameterization, dataset scale, or pretraining regime.
Autoregressive, diffusion, and flow-based models can all represent distributions over world-action trajectories.
A joint WAM can likewise be task-specific or pretrained on heterogeneous data.
Recent WAMs emphasize large-scale pretraining and generalization~\citep{zhu2025uwm,cen2025worldvla,ye2026dreamzero}.
Here, we focus on the control problem that comes with generating a joint trajectory.

\paragraph{Joint generation exposes an additional control reference.}
An action-only generator gives downstream control only an action trajectory.
A joint WAM also provides a future trajectory $\tilde Z$.
For this extra output to matter for control, it should affect the resulting actions rather than act only as an auxiliary prediction.
WATO uses $\tilde A$ and $\tilde Z$ as references in the trajectory projection $\Pi_{\theta,g}$, so both can shape the optimized action trajectory.
Proposition~1 in Sec.~\ref{sec:wato_diff} makes this explicit: the state reference acts through the world model rollout Jacobian.

\paragraph{Scaling does not affect the control objective.}
Learning a broad trajectory distribution and choosing good behavior for a downstream task are different problems.
A large joint WAM may represent diverse behaviors and futures, but imitation alone does not ensure that a generated trajectory minimizes the task objective or matches the dynamics used for control.
Pretraining determines what trajectory distribution the model represents.
WATO shapes how that distribution is used for control.

\paragraph{The formulation is independent of the generator parameterization.}
Our implementation uses flow matching, but WATO is defined at the level of generated world-action trajectories.
The operator $\Pi_{\theta,g}(\tilde A,\tilde Z)$ depends on action and state references and differentiable action-conditioned dynamics, not on the flow architecture itself.
The same idea may therefore extend to other differentiable joint generative models, including larger pretrained WAMs.
We leave this direction to future work.


\section{Why WATO Should Be Your Favorite Controller}
\label{app:wato_controller}

Our results suggest WATO as a \textcolor{pubblue}{\emph{default}} controller for visual offline goal-conditioned control with local goals or waypoints.
This raises a broader question: why does WATO also outperform value-based GCRL methods designed for long-horizon goal reaching?
We look at local versus long-horizon control, representation, candidate selection, and the way each method learns a controller.

\paragraph{Short horizons do not make policy learning trivial.}
OGBench uses goal relabeling during training and predefined start-goal tasks for evaluation~\citep{park2025ogbench}.
Our evaluation instead samples goals $25$ steps ahead in test trajectories and allows $50$ execution steps.
Each pair has an observed local connection, so little cross-trajectory stitching is needed.
But the controller still needs accurate contact, pose control, and recovery from execution errors.
Being able to propagate goal-reaching values over long horizons does not automatically give a strong local controller.

\paragraph{The gap is not explained by the encoder alone.}
The main comparisons use the same pretrained LeWM encoder to control for representation quality.
End-to-end training with the official IMPALA-based setup raises GCIQL's average success from $50.9\%$ to $53.9\%$, with larger gains on \texttt{scene-play} and \texttt{puzzle-3x3}.
So encoder training can help GCIQL, but it does not close the gap.

\paragraph{The advantage persists without candidate selection.}
WATO with $K=1$ achieves $88.7\%$ average success and outperforms both GCIQL variants in all $7$ environments.
Increasing $K$ to $8$ adds another $3.0$ percentage points.
With $K=1$, world model ranking cannot change the executed proposal, so best-of-$K$ selection alone cannot explain the advantage.
\Cref{tab:app_controller_controls} summarizes both controls.

\begin{table}[!htbp]
\centering
\caption{
\textbf{Encoder and candidate-count controls.}
Success rates (\%, mean $\pm$ std over $4$ seeds).
End-to-end GCIQL uses the official IMPALA-based setup.
}
\label{tab:app_controller_controls}
\scriptsize
\setlength{\tabcolsep}{7pt}
\begin{tabular}{lcccc}
\toprule
& \multicolumn{2}{c}{GCIQL}
& \multicolumn{2}{c}{WATO: pretrained LeWM} \\
\cmidrule(lr){2-3}
\cmidrule(lr){4-5}
Environment
& \shortstack{Pretrained\\LeWM}
& \shortstack{End-to-end\\training}
& $K=1$
& $K=8$ \\
\midrule
\texttt{TwoRooms}
& $80$ \std{5}
& $81$ \std{8}
& $90$ \std{3}
& $93$ \std{0} \\

\texttt{Reacher}
& $71$ \std{4}
& $67$ \std{6}
& $82$ \std{5}
& $85$ \std{2} \\

\texttt{PushT}
& $39$ \std{0}
& $38$ \std{5}
& $96$ \std{3}
& $100$ \std{0} \\

\texttt{cube-single}
& $52$ \std{1}
& $56$ \std{7}
& $97$ \std{2}
& $99$ \std{1} \\

\texttt{cube-double}
& $37$ \std{5}
& $40$ \std{10}
& $86$ \std{2}
& $90$ \std{3} \\

\texttt{scene-play}
& $40$ \std{4}
& $51$ \std{5}
& $90$ \std{1}
& $94$ \std{1} \\

\texttt{puzzle-3x3}
& $37$ \std{6}
& $44$ \std{4}
& $80$ \std{3}
& $81$ \std{2} \\
\midrule
Average
& $50.9$
& $\mathbf{53.9}$
& $\mathbf{88.7}$
& $91.7$ \\
\bottomrule
\end{tabular}
\end{table}

\paragraph{Trajectory supervision directly matches the local control problem.}
Each training window gives a current state, an attained future goal, and the actions and states in between.
Goal-conditioned imitation uses the same kind of hindsight supervision~\citep{ghosh2021gcsl}, and GCRL also uses relabeled goals and dataset actions.
So the key difference is not whether the methods see goal-reaching trajectories, but how they use them.
GCIQL learns goal-conditioned values and extracts a policy using those values and a behavioral constraint~\citep{park2025ogbench}.
WATO instead learns the joint trajectory and sends task gradients through its optimized solution.
In this sense, it trades bootstrapped value learning for model-based trajectory supervision while keeping the imitation signal.

Simply matching the goal distribution throughout trajectories is not enough.
GCBC and the flow policy achieve $66.4\%$ and $69.3\%$ average success, compared with $79.9\%$ for Joint WAF and $91.7\%$ for WATO.
Neither hindsight imitation nor action-only generation reaches the performance of WATO.

\paragraph{One trajectory sample can still represent coordinated behavior.}
Reducing $K$ removes best-of-$K$ selection, but not the structure within a sampled trajectory.
A joint WAF can generate an action sequence whose approach, contact, and subsequent motion follow one behavioral mode.
This may help compared with extracting individual actions from a unimodal policy, becasue the continuous-action OGBench baselines use Gaussian policies evaluated at their means~\citep{park2025ogbench}.
Still, our comparisons do not cleanly separate temporal coordination, policy expressiveness, and the training objective.
We therefore view trajectory generation as a useful inductive bias, not as something a Markov policy could never represent.

\paragraph{Lower GCRL performance does not point to value error alone.}
The bottleneck could lie in value estimation, policy extraction, or generalization to states encountered during execution.
Prior work shows that policy extraction and test-time generalization can limit offline RL even when value learning works well~\citep{park2024bottlenecks}.
Our controls do not isolate these factors one by one, but they show that neither encoder choice nor best-of-$K$ selection explains the gap.

WATO takes a different route: it gets continuous task supervision through a learned rollout and trajectory optimization.
This still depends on world model accuracy and on transferring optimized behavior into the unprojected generator, as analyzed in Appendix~\ref{app:optimization_amortization}.

\paragraph{WATO as the default local controller.}
Taken together, our results point to a simple default: if the goal is to turn visual offline trajectories into a goal-conditioned controller, start with a joint WAM and optimize it for control.
WATO outperforms the evaluated GCRL methods even when GCIQL trains its visual encoder end-to-end, and $K=1$ WATO remains far ahead without best-of-$K$ selection.
The gap is therefore not explained by representation quality or test-time candidate selection alone.
WATO also keeps the resulting controller simple at deployment, requiring only trajectory generation and a small number of world model evaluations.

This does not make WAMs and GCRL mutually exclusive.
Generative policies can be combined with value learning \citep{park2025fql}, and values may be useful when the problem requires trajectory stitching or reasoning far beyond locally supported trajectories.
But that is a different job from learning a strong local controller.
For the controller itself, our results give a clear answer: directly learn the trajectory, optimize it for the task, and amortize that optimization into the generator.

\begin{center}
\colorbox{corebg}{
\parbox{0.92\linewidth}{
\vspace{0.35em}
\centering
For visual offline goal-conditioned control, \textcolor{pubblue}{\textbf{\emph{WATO should be your favorite controller}}}.
\vspace{0.35em}
}}
\end{center}


\section{Theory and Derivations}
\label{app:theory}

This section develops the reference sensitivities and local geometry used in Sec.~\ref{sec:wato_diff}.
We also study the local stability of the trajectory projection and connect its residuals to optimization amortization.
All results are \emph{local}, since the world model rollout and trajectory optimization problem are generally nonlinear and nonconvex.

\subsection{Reference Sensitivity}
\label{app:reference_sensitivity}

We start from the reduced WATO objective
\begin{equation}
    \Psi(A;\tilde A,\tilde r)
    =
    F(A)
    +
    \frac{\lambda_A}{H}\|A-\tilde A\|_2^2
    +
    \frac{\lambda_Z}{H}\|r(A)-\tilde r\|_2^2,
    \label{eq:app_wato_reduced}
\end{equation}
where
$F(A)=J(\mathcal R_\theta(z_t,A),g)$,
$r(A)=\operatorname{vec}\mathcal N_Z(\mathcal R_\theta(z_t,A))$,
and
$\tilde r=\operatorname{vec}\mathcal N_Z(\tilde Z)$.
Let
$S(A)=\partial r(A)/\partial A$
and write the first-order condition as
\begin{equation}
    G(A;\tilde A,\tilde r)
    :=
    \nabla_A\Psi(A;\tilde A,\tilde r)
    =
    0.
    \label{eq:app_wato_stationarity}
\end{equation}

For an interior local solution $A^\star$, this becomes
\begin{equation}
    \nabla F(A^\star)
    +
    \frac{2\lambda_A}{H}(A^\star-\tilde A)
    +
    \frac{2\lambda_Z}{H}
    S_\star^\top
    \bigl(r(A^\star)-\tilde r\bigr)
    =
    0,
    \label{eq:app_wato_foc}
\end{equation}
where $S_\star=S(A^\star)$.
Let
$\mathcal H_\star
=
\nabla_{AA}^2
\Psi(A^\star;\tilde A,\tilde r)$.
If $\mathcal H_\star$ is nonsingular, the implicit function theorem gives
\begin{equation}
    \frac{\partial A^\star}{\partial y}
    =
    -
    \mathcal H_\star^{-1}
    \frac{\partial G}{\partial y},
    \qquad
    y\in\{\tilde A,\tilde r\}.
    \label{eq:app_ift}
\end{equation}

The action reference enters~\Cref{eq:app_wato_foc} directly, with
$\partial G/\partial\tilde A
=
-(2\lambda_A/H)I$.
This gives
\begin{equation}
    \frac{\partial A^\star}{\partial\tilde A}
    =
    \frac{2\lambda_A}{H}
    \mathcal H_\star^{-1}.
    \label{eq:app_action_sensitivity}
\end{equation}

The state reference enters through the world model rollout.
Since
$\partial G/\partial\tilde r
=
-(2\lambda_Z/H)S_\star^\top$,
we get
\begin{equation}
    \frac{\partial A^\star}{\partial\tilde r}
    =
    \frac{2\lambda_Z}{H}
    \mathcal H_\star^{-1}S_\star^\top.
    \label{eq:app_state_sensitivity}
\end{equation}

Together, \Cref{eq:app_action_sensitivity}-\Cref{eq:app_state_sensitivity} give Proposition~1.
The action reference changes the optimized action trajectory through the local optimization Hessian, while the state reference additionally passes through $S_\star^\top$, which maps changes in the predicted future back to action space.
For box-constrained solutions, the same derivation applies locally on the free action coordinates when the active set is fixed.

\subsection{Local Optimization Geometry}
\label{app:local_geometry}

The Hessian of \Cref{eq:app_wato_reduced} also shows how the two references shape the local optimization geometry.
Define the state-reference residual $e(A)=r(A)-\tilde r$.
Holding $\tilde r$ fixed, its squared norm has gradient
\begin{equation}
    \nabla_A \|e(A)\|_2^2
    =
    2S(A)^\top e(A).
    \label{eq:app_state_ref_gradient}
\end{equation}
Writing $r_i(A)$ and $e_i(A)$ for the corresponding coordinates and differentiating again gives
\begin{equation}
    \nabla_A^2 \|e(A)\|_2^2
    =
    2S(A)^\top S(A)
    +
    2\sum_i
    e_i(A)\nabla_A^2 r_i(A).
    \label{eq:app_state_ref_hessian}
\end{equation}
The exact Hessian at $A^\star$ is
\begin{equation}
    \mathcal H_\star
    =
    \nabla^2 F(A^\star)
    +
    \frac{2\lambda_A}{H}I
    +
    \frac{2\lambda_Z}{H}
    \left[
        S_\star^\top S_\star
        +
        \sum_i
        e_{\star,i}\nabla_A^2 r_i(A^\star)
    \right],
    \label{eq:app_exact_hessian}
\end{equation}
where $e_\star=r(A^\star)-\tilde r$.

Approximating the state-reference Hessian by its Gauss-Newton term gives
\begin{equation}
    \mathcal H_\star
    \approx
    \nabla^2 F(A^\star)
    +
    \frac{2\lambda_A}{H}I
    +
    \frac{2\lambda_Z}{H}S_\star^\top S_\star.
    \label{eq:app_gn_hessian}
\end{equation}
The task Hessian $\nabla^2 F(A^\star)$ is kept exact.
If $\|\nabla_A^2 r_i(A)\|\le M_r$ for every coordinate in a neighborhood of $A^\star$, the omitted term satisfies
\begin{equation}
    \left\|
    \frac{2\lambda_Z}{H}
    \sum_i
    e_{\star,i}\nabla_A^2 r_i(A^\star)
    \right\|
    \le
    \frac{2\lambda_Z M_r}{H}
    \|e_\star\|_1.
    \label{eq:app_geometry_residual_bound}
\end{equation}
So the approximation error vanishes with the state-reference residual.

The action reference contributes isotropic curvature $(2\lambda_A/H)I$.
The state-reference Gauss-Newton term is positive semidefinite since, for any action perturbation $u$,
\begin{equation}
    u^\top S_\star^\top S_\star u
    =
    \|S_\star u\|_2^2
    \ge
    0.
    \label{eq:app_psd_geometry}
\end{equation}
It weights action directions by how strongly they change the standardized predicted trajectory.
The task Hessian and residual-weighted term may still be indefinite, so this does not make the full objective convex.

\paragraph{Generated versus rollout-derived references.}
One interesting distinction is that this geometry does not require a WAM-generated future.
For a fixed proposal $\tilde A$, replacing $\tilde r$ with $r(\tilde A)$ gives the state-reference penalty
$\lambda_Z H^{-1}\|r(A)-r(\tilde A)\|_2^2$.
At the same action trajectory $A$, its Gauss-Newton contribution is still $(2\lambda_Z/H)S(A)^\top S(A)$.
What changes is the reference itself, and hence the gradient, residual-weighted Hessian term, and optimized solution.
Removing $d_Z$ therefore tests the state-reference penalty as a whole, but does not separate generated-future information from dynamics-aware regularization.
This motivates the generated versus rollout-derived reference comparison in Appendix~\ref{app:full_wato_ablations}.

\subsection{Projection Stability}
\label{app:projection_stability}

The same sensitivity expressions also tell us how stable the projection is locally.
Assume that $\Psi$ is locally $\mu$-strongly convex in $A$ around $A^\star$, so that
\begin{equation}
    \mathcal H_\star
    \succeq
    \mu I,
    \qquad
    \mu>0.
    \label{eq:app_strong_convexity}
\end{equation}
Then
$\|\mathcal H_\star^{-1}\|\le 1/\mu$.
Using~\Cref{eq:app_action_sensitivity},
\begin{equation}
    \left\|
    \frac{\partial A^\star}{\partial\tilde A}
    \right\|
    \le
    \frac{2\lambda_A}{H\mu}.
    \label{eq:app_action_stability}
\end{equation}
Similarly,
\begin{equation}
    \left\|
    \frac{\partial A^\star}{\partial\tilde r}
    \right\|
    \le
    \frac{2\lambda_Z}{H\mu}
    \|S_\star\|.
    \label{eq:app_state_stability}
\end{equation}

For sufficiently small perturbations
$(\Delta\tilde A,\Delta\tilde r)$,
these bounds give
\begin{equation}
    \|\Delta A^\star\|
    \le
    \frac{2}{H\mu}
    \left(
        \lambda_A\|\Delta\tilde A\|
        +
        \lambda_Z\|S_\star\|
        \|\Delta\tilde r\|
    \right)
    +
    o(\|(\Delta\tilde A,\Delta\tilde r)\|).
    \label{eq:app_projection_lipschitz}
\end{equation}

The action-reference sensitivity is controlled by the local curvature of the optimization problem.
For the state reference, the rollout Jacobian also matters: it has more influence in directions where small action changes produce larger changes in the future.

\subsection{Optimization Amortization}
\label{app:optimization_amortization}

There is one subtlety in WATO: training evaluates the projected task cost $F(A^\star)$, while deployment uses the unprojected proposal $\tilde A$.
A good projected trajectory does not automatically imply a good raw proposal.
For example, consider $H=1$, $r(A)=A$, $F(A)=A^2$, and $\lambda_A=\lambda_Z=1$.
The projection is
\begin{equation}
    A^\star
    =
    \operatorname*{arg\,min}_A
    \left[
        A^2+(A-\tilde A)^2+(A-\tilde r)^2
    \right]
    =
    \frac{\tilde A+\tilde r}{3}.
    \label{eq:app_amortization_counterexample}
\end{equation}
Setting $\tilde A=t$ and $\tilde r=-t$ gives $F(A^\star)=0$ but $F(\tilde A)=t^2$.
This example only concerns the projected task term.
WATO still keeps trajectory imitation through $\mathcal L_{\mathrm{WAF}}$.
The useful question is therefore when a small projection residual is enough to transfer the optimized behavior back to the raw proposal.

\paragraph{Residual-based deployment bounds.}
For brevity, write
$d_A=d_A(A^\star,\tilde A)$
and
$d_Z=d_Z(Z^\star,\tilde Z)$,
where $Z^\star=\mathcal R_\theta(z_t,A^\star)$.
By definition,
\begin{equation}
    \|A^\star-\tilde A\|_2
    =
    \sqrt{H d_A},
    \qquad
    \|r(A^\star)-\tilde r\|_2
    =
    \sqrt{H d_Z}.
    \label{eq:app_projection_residuals}
\end{equation}
Suppose $A^\star$ is an exact interior local solution.
Assume that $\nabla F$ is $L_F$-Lipschitz and $r$ is $L_r$-Lipschitz on a neighborhood containing the segment between $A^\star$ and $\tilde A$.
Below, matrix norms are spectral norms.

\medskip
\noindent
\textcolor{pubblue}{\textbf{Residual-based deployment transfer.}}
\vspace{0.25em}

\noindent
\colorbox{corebg}{%
\parbox{\dimexpr\linewidth-2\fboxsep\relax}{
The task-cost gap satisfies
\begin{equation}
    \left|F(\tilde A)-F(A^\star)\right|
    \le
    \left(
        2\lambda_A+\frac{H L_F}{2}
    \right)d_A
    +
    2\lambda_Z\|S_\star\|\sqrt{d_A d_Z}.
    \label{eq:app_deployment_cost_bound}
\end{equation}
The inconsistency between the generated future and the rollout of the generated actions satisfies
\begin{equation}
    \sqrt{
        d_Z\!\left(
            \mathcal R_\theta(z_t,\tilde A),
            \tilde Z
        \right)
    }
    \le
    L_r\sqrt{d_A}
    +
    \sqrt{d_Z}.
    \label{eq:app_raw_consistency_bound}
\end{equation}
}}
\medskip

\emph{Proof.}
The first-order condition in \Cref{eq:app_wato_foc} gives
\begin{equation}
    \|\nabla F(A^\star)\|_2
    \le
    \frac{2}{\sqrt{H}}
    \left(
        \lambda_A\sqrt{d_A}
        +
        \lambda_Z\|S_\star\|\sqrt{d_Z}
    \right).
    \label{eq:app_amortization_bound}
\end{equation}
Lipschitz continuity of $\nabla F$ gives
\begin{equation}
    \left|F(\tilde A)-F(A^\star)\right|
    \le
    \|\nabla F(A^\star)\|_2\sqrt{H d_A}
    +
    \frac{H L_F}{2}d_A.
    \label{eq:app_task_cost_expansion}
\end{equation}
Substituting \Cref{eq:app_amortization_bound} proves \Cref{eq:app_deployment_cost_bound}.
For the consistency bound, the triangle inequality gives
\begin{equation}
    \|r(\tilde A)-\tilde r\|_2
    \le
    L_r\|\tilde A-A^\star\|_2
    +
    \|r(A^\star)-\tilde r\|_2.
    \label{eq:app_consistency_triangle}
\end{equation}
Dividing by $\sqrt{H}$ gives \Cref{eq:app_raw_consistency_bound}.

These bounds connect the reference distances in \Cref{fig:wato_correction} to two deployment quantities: the cost of removing optimization and the world model inconsistency of the generated trajectory.
The second statement is with respect to the pretrained world model, not the true environment.

\paragraph{Stationarity and fixed points.}
Under the same assumptions,
\begin{equation}
    \|\nabla F(\tilde A)\|_2
    \le
    \frac{2}{\sqrt{H}}
    \left[
        \left(
            \lambda_A+\frac{H L_F}{2}
        \right)\sqrt{d_A}
        +
        \lambda_Z\|S_\star\|\sqrt{d_Z}
    \right].
    \label{eq:app_proposal_stationarity}
\end{equation}
This follows from
$\|\nabla F(\tilde A)\|_2
\le
\|\nabla F(A^\star)\|_2
+
L_F\|\tilde A-A^\star\|_2$
and \Cref{eq:app_amortization_bound}.
So small residuals also bound the task gradient of the raw proposal.

If $A^\star=\tilde A$ and $\tilde r=r(\tilde A)$, both distances vanish and
\begin{equation}
    \nabla_A
    J(\mathcal R_\theta(z_t,\tilde A),g)
    =
    0.
    \label{eq:app_fixed_point_stationarity}
\end{equation}
An interior dynamics-consistent fixed point is task-stationary and not necessarily globally optimal.

\paragraph{Active action bounds.}
When action bounds are active, the full task gradient need not vanish.
Let $N_{\mathcal U^H}(A)$ denote the normal cone of the action box.
Constrained optimality gives
$0\in\nabla_A\Psi(A^\star;\tilde A,\tilde r)+N_{\mathcal U^H}(A^\star)$,
which implies
\begin{equation}
    \operatorname{dist}
    \left(
        0,\,
        \nabla F(A^\star)+N_{\mathcal U^H}(A^\star)
    \right)
    \le
    \frac{2}{\sqrt{H}}
    \left(
        \lambda_A\sqrt{d_A}
        +
        \lambda_Z\|S_\star\|\sqrt{d_Z}
    \right).
    \label{eq:app_constrained_stationarity}
\end{equation}
At a dynamics-consistent fixed point, this becomes constrained task stationarity.

For feasible $\tilde A$ and $A^\star$ in a neighborhood where $F$ is $C_F$-Lipschitz, we can still bound the task-cost gap without assuming interior stationarity:
\begin{equation}
    \left|F(\tilde A)-F(A^\star)\right|
    \le
    C_F\sqrt{H d_A}.
    \label{eq:app_constrained_cost_bound}
\end{equation}
The dynamics-consistency bound in \Cref{eq:app_raw_consistency_bound} also continues to hold at active bounds, since its proof does not use stationarity.

\paragraph{Inexact optimization.}
The same picture extends to approximate solves.
For an approximate interior solution, let
$\epsilon_{\mathrm{opt}}
=
\|\nabla_A\Psi(A^\star;\tilde A,\tilde r)\|_2$,
with all distances and Jacobians evaluated at the returned trajectory.
The right-hand sides of \Cref{eq:app_amortization_bound,eq:app_proposal_stationarity} then gain $\epsilon_{\mathrm{opt}}$, while \Cref{eq:app_deployment_cost_bound} gains $\epsilon_{\mathrm{opt}}\sqrt{H d_A}$.
For a bound-active solution, \Cref{eq:app_constrained_stationarity} similarly gains the corresponding KKT residual
$\operatorname{dist}(0,\nabla_A\Psi(A^\star;\tilde A,\tilde r)+N_{\mathcal U^H}(A^\star))$.
The consistency and Lipschitz cost bounds do not require a converged projection.

\paragraph{Candidate selection under model error.}
Finally, consider the deployed best-of-$K$ rule.
For $K$ generated candidates and their projected trajectories, let
$k^\star=\operatorname*{arg\,min}_k F(\tilde A^{(k)})$.
Let $F_{\mathrm{env}}(A)$ denote the terminal task cost obtained by executing $A$ in the environment, using the same encoder and goal.
Suppose
$|F_{\mathrm{env}}(A)-F(A)|\le\epsilon_{\mathrm{model}}$
for every generated and projected action trajectory in this candidate set.
Let $\eta_k\ge0$ satisfy
$F(\tilde A^{(k)})\le F(A^{\star(k)})+\eta_k$,
as provided by the preceding task-cost bounds.
Then
\begin{equation}
    \begin{aligned}
        F_{\mathrm{env}}(\tilde A^{(k^\star)})
        &\le
        \min_k
        \left[
            F_{\mathrm{env}}(A^{\star(k)})+\eta_k
        \right]
        +
        2\epsilon_{\mathrm{model}}
        \\
        &\le
        \min_k F_{\mathrm{env}}(A^{\star(k)})
        +
        \max_k\eta_k
        +
        2\epsilon_{\mathrm{model}}.
    \end{aligned}
    \label{eq:app_selection_transfer}
\end{equation}
This follows from
$F(\tilde A^{(k^\star)})
\le
F(\tilde A^{(k)})
\le
F(A^{\star(k)})+\eta_k$
and the model-error bound on the selected and projected trajectories.
The bound separates three effects: projected candidate quality, the cost of removing optimization, and world model error.

These results describe what small projection residuals buy us: WATO does not explicitly minimize them.
They are finite-horizon statements and do not imply convergence of WATO training, closed-loop stability, or environment-defined success.


\section{Trajectory Optimization and Differentiation}
\label{app:trajectory_optimization}

WATO optimizes only the action trajectory and obtains future states from the world model rollout.
The world model parameters remain fixed, while derivatives w.r.t. the action trajectory are retained.

\subsection{Reduced-Space Optimization}
\label{app:reduced_space_optimization}

We implement~\Cref{eq:wato_trajectory_optimization} in standardized action coordinates.
The action trajectory $A$ is the only optimization variable, with
$Z(A)=\mathcal R_\theta(z_t,A)$.
We solve
\begin{equation}
    A^\star
    =
    \operatorname*{arg\,min}_{A\in\mathcal U_{\mathrm{std}}^H}
    J(Z(A),g)
    +
    \lambda_A d_A(A,\tilde A)
    +
    \lambda_Z d_Z(Z(A),\tilde Z),
    \qquad
    Z^\star=Z(A^\star),
    \label{eq:app_reduced_optimization}
\end{equation}
where $\mathcal U_{\mathrm{std}}$ denotes the action bounds in standardized coordinates.
The optimized future $Z^\star$ is always produced by the world model.

We use
\begin{equation}
    J(Z(A),g)
    =
    \|Z_H(A)-g\|_2^2,
    \label{eq:app_task_cost}
\end{equation}
and
\begin{equation}
    d_A(A,\tilde A)
    =
    \frac{1}{H}
    \sum_{h=0}^{H-1}
    \|A_h-\tilde A_h\|_2^2,
    \qquad
    d_Z(Z,\tilde Z)
    =
    \frac{1}{H}
    \sum_{h=1}^{H}
    \|
        \mathcal N_Z(z_h)
        -
        \mathcal N_Z(\tilde z_h)
    \|_2^2.
    \label{eq:app_reference_distances}
\end{equation}

Primitive actions are standardized using offline-data statistics before forming action blocks.
Environment action bounds are transformed with the same statistics and imposed as box constraints.
The action reference $\tilde A$ uses the same coordinates.

The state normalization is
$\mathcal N_Z(z)=(z-\mu_Z)/\sigma_Z$,
where $\mu_Z$ and $\sigma_Z$ are computed only from frozen-LeWM latents in the offline training split.
This normalization is used only in $d_Z$; the task cost is evaluated in the native pretrained latent space.
We use $\lambda_A=10$ and $\lambda_Z=0.1$ in all experiments.

\subsection{Numerical Solver}
\label{app:numerical_solver}

We initialize the primal solve from the clipped generated action reference $\tilde A$.
We run L-BFGS-B for at most $200$ iterations under the standardized action bounds, followed by up to $25$ projected-gradient polishing iterations.
Each polishing iteration uses at most $12$ backtracking trials.

Because the world model rollout is nonlinear, the optimization problem is generally nonconvex.
We require local stationarity using
\begin{equation}
    \left\|
        \nabla_{\mathrm{proj}}
        \Psi(A^\star;\tilde A,\tilde r)
    \right\|_\infty
    \le
    0.02,
    \label{eq:app_stationarity_check}
\end{equation}
where the projected gradient is taken with respect to the standardized action bounds.

During training, a task sample whose solution fails~\Cref{eq:app_stationarity_check} is rejected and retried with another reserved window and proposal.
We allow at most four attempts per update.
If none satisfies the stationarity criterion, the task update is aborted.

The trajectory optimizer is used only during WATO training.
Deployment uses generated proposals and world model evaluation without L-BFGS-B refinement.

\subsection{Implicit Differentiation}
\label{app:implicit_diff}

We differentiate the local solution through its optimality conditions rather than unrolling the primal solver.
Let
$y=(\tilde A,\tilde r)$,
where
$\tilde r=\operatorname{vec}\mathcal N_Z(\tilde Z)$,
and let $A^\star(y)$ denote a local solution.
For an interior regular solution,
$\nabla_A\Psi(A^\star;y)=0$.
Define
$\mathsf H=\nabla_{AA}^2\Psi(A^\star;y)$
and
$\mathsf B=\partial_y\nabla_A\Psi(A^\star;y)$.
Differentiating the stationarity condition gives
\begin{equation}
    \mathsf H
    \frac{\partial A^\star}{\partial y}
    +
    \mathsf B
    =
    0.
    \label{eq:app_implicit_system}
\end{equation}

We only need the vector-Jacobian product with the task loss.
Let
\begin{equation}
    g_{\mathrm{task}}
    =
    \left.
    \nabla_A
    J(\mathcal R_\theta(z_t,A),g)
    \right|_{A=A^\star}.
    \label{eq:app_task_action_gradient}
\end{equation}
Rather than forming $\mathsf H^{-1}$, we solve
\begin{equation}
    \mathsf H^\top v
    =
    g_{\mathrm{task}},
    \qquad
    \nabla_y J(Z^\star,g)
    =
    -\mathsf B^\top v.
    \label{eq:app_implicit_adjoint}
\end{equation}
The resulting reference gradients are propagated through the joint generator to update $\phi$.

Let
\begin{equation}
    J_R
    =
    \left.
    \frac{\partial\mathcal R_\theta(z_t,A)}
         {\partial A}
    \right|_{A=A^\star},
\end{equation}
and let $D_Z$ denote the block-diagonal linear map induced by $\mathcal N_Z$.
The standardized rollout Jacobian is
$S_\star=D_ZJ_R$.
The joint WAF directly outputs the normalized state reference $\tilde r$, with task gradient
\begin{equation}
    \nabla_{\tilde A}J(Z^\star,g)
    =
    \frac{2\lambda_A}{H}v,
    \qquad
    \nabla_{\tilde r}J(Z^\star,g)
    =
    \frac{2\lambda_Z}{H}
    D_Z J_R v.
    \label{eq:app_normalized_reference_gradients}
\end{equation}
In native latent coordinates,
\begin{equation}
    \nabla_{\tilde Z}J(Z^\star,g)
    =
    D_Z^\top
    \nabla_{\tilde r}J(Z^\star,g)
    =
    \frac{2\lambda_Z}{H}
    D_Z^\top D_Z J_R v.
    \label{eq:app_reference_gradients}
\end{equation}
Thus, the generated future receives a task gradient through its effect on the optimized actions even though it is not an independent optimization variable.

For a box-constrained solution with a locally fixed active set, we differentiate only through the free action coordinates.
Let $E_{\mathrm{free}}$ embed the free coordinates into the full action vector.
The theoretical adjoint system is
\begin{equation}
    \left(
        E_{\mathrm{free}}^\top
        \mathsf H
        E_{\mathrm{free}}
    \right)^\top
    v_{\mathrm{free}}
    =
    E_{\mathrm{free}}^\top
    g_{\mathrm{task}},
    \qquad
    v
    =
    E_{\mathrm{free}}v_{\mathrm{free}}.
    \label{eq:app_active_set_adjoint}
\end{equation}
This local derivative does not cover points where the active set changes.

In implementation, we solve the damped system
\begin{equation}
    \left(
        E_{\mathrm{free}}^\top
        \mathsf H
        E_{\mathrm{free}}
        +
        10^{-3}I
    \right)^\top
    v_{\mathrm{free}}
    =
    E_{\mathrm{free}}^\top
    g_{\mathrm{task}}.
    \label{eq:app_damped_adjoint}
\end{equation}
We compute Hessianvector products through the frozen world model and use conjugate gradients for at most $16$ iterations with tolerance $10^{-4}$.
If conjugate gradients fails, we fall back to a dense solve and, when needed, a pseudoinverse.
Damping and finite solver tolerance make the backward pass an approximation to the local sensitivity.

The world model parameters $\theta$ are never updated by this gradient.
Derivatives through $\mathcal R_\theta$ with respect to the action trajectory are retained.


\section{World Model and WAF Implementation}
\label{app:wm_waf}

We specify the pretrained world model, joint world-action generator, and WATO training procedure.
The world model and joint WAF are trained before WATO.
WATO updates only the generator parameters $\phi$ and the world model parameters $\theta$ remain fixed.

\subsection{World Model}
\label{app:world_model}

We use latent world models following LeWM~\citep{maes2026lewm}.
Each model consists of a visual encoder $E_\theta$ and an action-conditioned predictor $F_\theta$ trained jointly end-to-end.
The encoder maps an observation to a latent state, while the predictor takes an action block $A_h$ and a finite latent context $\mathcal C_h$ and predicts the next latent state.
Autoregressive prediction gives the rollout $\mathcal R_\theta$ defined in \Cref{eq:latent_rollout}.

The released LeWM architecture uses $224\times224$ visual inputs, a ViT-Tiny encoder with patch size $14$, and a six-layer action-conditioned Transformer predictor.
For released checkpoints, five primitive actions are grouped into each action block.
The predictor uses a finite latent context whose length depends on the environment.

LeWM is trained with next-latent prediction and SIGReg.
The prediction loss is
\begin{equation}
    \mathcal L_{\mathrm{pred}}
    =
    \mathbb E_{\mathcal D}
    \left[
        \operatorname{MSE}
        \left(
            \hat z_{h+1},
            z_{h+1}
        \right)
    \right],
    \label{eq:app_world_model_prediction}
\end{equation}
and the full objective is
$\mathcal L_{\mathrm{WM}}
=
\mathcal L_{\mathrm{pred}}
+
\lambda_{\mathrm{sig}}
\mathcal L_{\mathrm{SIG}}$.
Following LeWM, we use $1024$ random projections for SIGReg and $\lambda_{\mathrm{sig}}=0.1$.

We use released LeWM checkpoints when available.
For environments without one, we train the same LeWM architecture and objective from scratch.
We retain the default temporal abstraction of five primitive actions per latent transition and a maximum context of three latent states.
For models trained in this work, the training entry reports the optimizer step of the selected validation checkpoint.
\Cref{tab:world_model_configs} summarizes the configurations.

\begin{table}[!htbp]
\centering
\caption{
\textbf{World model configurations.}
$B$ is the number of primitive actions per latent transition and $l_c$ is
the maximum latent context length.
}
\label{tab:world_model_configs}
\scriptsize
\setlength{\tabcolsep}{4pt}
\begin{tabular}{lcccc}
\toprule
Environment & Source & $B$ & $l_c$ & Selected checkpoint \\
\midrule
\texttt{TwoRooms}
    & Released LeWM
    & $5$
    & $1$
    & Released \\

\texttt{Reacher}
    & Released LeWM
    & $5$
    & $3$
    & Released \\

\texttt{PushT}
    & Released LeWM
    & $5$
    & $3$
    & Released \\

\texttt{cube-single}
    & Released LeWM
    & $5$
    & $3$
    & Released \\

\texttt{cube-double}
    & Trained in this work
    & $5$
    & $3$
    & $92{,}052$ updates \\

\texttt{scene-play}
    & Trained in this work
    & $5$
    & $3$
    & $76{,}252$ updates \\

\texttt{puzzle-3x3}
    & Trained in this work
    & $5$
    & $3$
    & $68{,}724$ updates \\
\bottomrule
\end{tabular}
\end{table}

For world models trained in this work, we use AdamW with a learning rate of $5\times10^{-5}$, weight decay of $10^{-3}$, and batch size $128$.
We train for up to $12$ epochs (approximately $92$k optimizer updates) with a $1\%$ linear warmup followed by cosine decay to zero, and clip the global gradient norm to $1.0$.
We select the checkpoint with the lowest teacher-forced validation loss, defined as prediction MSE plus $0.09$ times the SIGReg loss.

All world models are trained before WATO.
During WATO, gradients through $E_\theta$, $F_\theta$, and $\mathcal R_\theta$ with respect to their inputs are retained, but $\theta$ is never updated.

\subsection{Joint World Action Flow}
\label{app:joint_waf}

The joint WAF models
$X=(A,Z)$,
where
$A=(A_0,\ldots,A_{H-1})$
and
$Z=(z_1,\ldots,z_H)$,
conditioned on the current and goal latents $c=(z_t,g)$.
A single conditional flow generates the action and future-state trajectories jointly rather than autoregressively over the horizon.

The vector field $v_\phi$ is parameterized by a Transformer with model width $256$, $4$ Transformer layers, $8$ attention heads, feed-forward dimension $1024$, no dropout, and a $64$-dimensional flow-time embedding.
Action and latent variables are projected to the common model width, processed jointly, and mapped back to their respective output spaces.
The model has approximately $3.5$M parameters.

Primitive actions are standardized before action blocks are formed.
Future state targets are obtained by encoding dataset observations with the pretrained encoder.
For each minibatch, we sample a horizon-$H$ trajectory window, form $X=(A,Z)$, and train with the joint WAF objective in \Cref{eq:wato_fm}.
We use $\lambda_{\mathrm{WAF}}=1$, batch size $512$, and $50{,}000$ pretraining updates.

At generation time, Gaussian noise with the same structure as $X$ is integrated from flow time $0$ to $1$.
We use Euler integration with NFE $=8$.
All action and state variables are updated jointly at each flow step.
Unless otherwise stated, deployment draws $K=8$ independent candidates.

\subsection{WATO Training}
\label{app:wato_training}

WATO fine-tunes the pretrained joint WAF using the objective in
\Cref{eq:wato_loss}.
The flow matching term preserves trajectory imitation, while the task term differentiates through the trajectory-optimization layer.
Only the WAF parameters $\phi$ are updated. The world model remains fixed.

We use AdamW with a learning rate of $5\times10^{-5}$, weight decay
$10^{-4}$, and a flow matching batch size of $64$ for $10{,}000$ WATO updates.
The task term uses one projected trajectory per update with
$\lambda_{\mathrm{task}}=5\times10^{-3}$.
For each update, trajectories for the task term are generated with Euler integration using NFE $=2$ and passed through the trajectory projection in \Cref{eq:wato_projection}.
We draw up to four proposals to obtain a stationarity-qualified trajectory.
Only solutions satisfying the criterion in
\Cref{eq:app_stationarity_check} contribute to the task gradient.

We clip the global generator gradient norm to $1$.
The learning rate uses a linear warmup over the first $100$ updates followed by cosine decay to zero.
We select the final checkpoint using the minimum fixed-validation objective,
$\mathcal{L}_{\mathrm{WAF}}+ \lambda_{\mathrm{task}}\mathcal{L}_{\mathrm{task}}$.

\begin{table}[!htbp]
\centering
\caption{
\textbf{Joint WAF training and WATO training configuration.}
NFE $=2$ is used to generate trajectories for the task loss during training, whereas all controllers use NFE $=8$ at deployment.
}
\label{tab:wato_training_config}
\scriptsize
\setlength{\tabcolsep}{5pt}
\begin{tabular}{lcc}
\toprule
Setting & Joint WAF training & WATO training \\
\midrule
Optimizer
    & AdamW
    & AdamW \\
Weight decay
    & $10^{-4}$
    & $10^{-4}$ \\
Learning rate
    & $3\times10^{-4}$
    & $5\times10^{-5}$ \\
Batch size
    & $512$
    & $64$ (FM), $1$ (task) \\
Updates
    & $50{,}000$
    & $10{,}000$ \\
Warmup
    & $500$ updates
    & $100$ updates \\
Learning-rate schedule
    & Cosine decay
    & Cosine decay \\
Task-loss flow integration
    & --
    & Euler, NFE $=2$ \\
Deployment flow integration
    & Euler, NFE $=8$
    & Euler, NFE $=8$ \\
$\lambda_{\mathrm{WAF}}$
    & $1$
    & $1$ \\
$\lambda_{\mathrm{task}}$
    & --
    & $5\times10^{-3}$ \\
Gradient-norm clipping
    & $1$
    & $1$ \\
Checkpoint criterion
    & Minimum validation FM loss
    & Minimum validation weighted objective \\
\bottomrule
\end{tabular}
\end{table}

\subsection{Training Cost and Solver Statistics}
\label{app:training_cost}

WATO adds training-time computation because each task update solves a trajectory optimization problem and differentiates through its solution.
\Cref{tab:wato_training_cost} reports statistics from the full $10{,}000$-update \texttt{PushT} run.
All wall-clock times are measured on a single NVIDIA RTX Pro 6000 Blackwell GPU.

WATO fine-tuning takes $5.82$ hours, compared with $1.76$ hours for joint WAF pretraining.
The primal solver uses $90$ L-BFGS-B iterations on average, but the stationarity filter rarely rejects a solve: only $5$ of $10{,}005$ attempts are rejected, and no training update is aborted.
The additional computation is confined to training; deployment uses generation, world model evaluation, and best-of-$K$ selection without iterative trajectory refinement.

\begin{table}[t]
\centering
\caption{
\textbf{Training cost and solver statistics on \texttt{PushT}.}
Statistics are from the full $10{,}000$-update WATO run on a single GPU.
}
\label{tab:wato_training_cost}
\scriptsize
\setlength{\tabcolsep}{3.5pt}

\begin{minipage}[h]{0.32\linewidth}
\centering
\vspace{0.5em}
\textbf{Training cost}

\vspace{0.5em}

\begin{tabular}{lc}
\toprule
Metric & Value \\
\midrule
Joint WAF pretraining
& $1.76$ h \\

WATO training
& $5.82$ h \\

Total pipeline
& $7.58$ h \\

WATO / Joint WAF
& $3.31\times$ \\

Pipeline / Joint WAF-only
& $4.31\times$ \\

FM-only continuation
& $0.65$ h \\

WATO / FM-only
& $9.0\times$ \\

Direct rollout
& $0.9$--$1.5$ h \\

WATO / direct rollout
& $3.9$--$6.5\times$ \\
\bottomrule
\end{tabular}
\end{minipage}
\hfill
\begin{minipage}[h]{0.32\linewidth}
\centering
\vspace{0.5em}
\textbf{Solver workload}

\vspace{0.65em}

\begin{tabular}{lc}
\toprule
Metric & Value \\
\midrule
Mean L-BFGS-B iter.
& $90$ \\

Median L-BFGS-B iter.
& $84$ \\

Interquartile range
& $53$--$116$ \\

Maximum L-BFGS-B iter.
& $200$ \\

Maximum polish iter.
& $25$ \\

Total L-BFGS-B iter.
& $8.8\times10^{5}$ \\
\bottomrule
\end{tabular}
\end{minipage}
\hfill
\begin{minipage}[h]{0.32\linewidth}
\centering
\vspace{0.5em}
\textbf{Solver reliability}

\vspace{0.45em}

\begin{tabular}{lc}
\toprule
Metric & Value \\
\midrule
Optimizer updates
& $10{,}000$ \\

Task-solve attempts
& $10{,}005$ \\

Rejected samples
& $5$ \\

First-attempt success
& $99.95\%$ \\

Success within $\leq4$
& $100\%$ \\

Aborted updates
& $0$ \\

Aborted-update rate
& $0\%$ \\
\bottomrule
\end{tabular}
\end{minipage}

\end{table}


\section{Experimental Protocol}
\label{app:experimental_protocol}

\begin{figure}[t]
    \centering
    \includegraphics[width=1.0\linewidth]{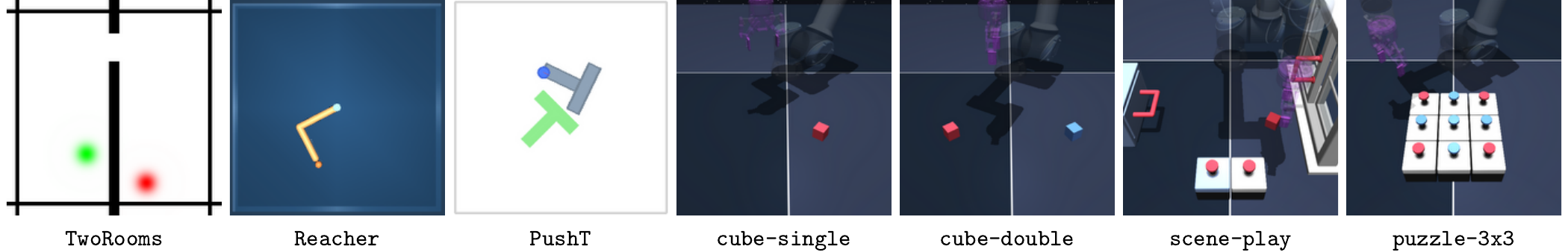}
    \caption{Environments for WATO experiments}
    \label{fig:environments}
\end{figure}

This section specifies goal construction, baseline training, environment-specific configurations, and compute measurement.
Unless stated otherwise, all methods use the same offline data, evaluation goals, execution budget, and pretrained visual representation.
\Cref{fig:environments} shows representative observations from the seven evaluation environments.

\subsection{Goal Construction and Evaluation}
\label{app:goal_evaluation}

We evaluate local visual goal reaching from evaluation trajectories.
For an initial observation $o_t$, the goal is the observation $o_{t+L}$ from the same trajectory, with $L=25$ environment steps.
All evaluation episodes are \emph{nontrivial}: the initial state never satisfies the goal criterion.
Only pairs with at least $L$ remaining steps are used.
This defines the same goal distribution across all methods.

Each episode has a $50$-step execution budget.
A controller repeatedly observes the current image, conditions on the fixed goal image, executes its selected action segment, and replans until success or the budget is exhausted.
All methods use the same environment-specific success criterion.

Success rate is the primary metric.
We also report the final latent goal error
$\|E_\theta(o_T)-g\|_2$,
where $o_T$ is the final observation.
All results use $4$ seeds and identical start-goal pairs across methods.

\subsection{Baseline Training}
\label{app:baseline_training}

All baselines are trained on the same offline dataset and goal distribution as WATO.
Evaluation uses identical start-goal pairs and the same execution budget.
All methods use the same pretrained LeWM encoder to separate control performance from representation quality.
Each baseline otherwise keeps its original objective, goal-relabeling procedure, and training scheme.

\paragraph{Goal-conditioned learning.}
GCBC~\citep{ding2019gcbc} learns a goal-conditioned policy by behavioral cloning.
GCIVL~\citep{kostrikov2021gcivl} and GCIQL~\citep{park2025ogbench} learn goal-conditioned value functions and policies from the same offline data.
QRL~\citep{wang2023qrl} learns goal-conditioned control through its quasimetric objective.
All four methods are trained from scratch for this goal distribution rather than evaluated with published final-goal checkpoints.

\paragraph{Generative control.}
The Flow policy generates only the action trajectory $A$ and serves as the action-only generative baseline.
Joint WAF uses the joint generator $q_\phi(A,Z\mid z_t,g)$ trained only with the flow matching objective in \Cref{eq:wato_fm}.
It isolates the effect of WATO training from joint trajectory imitation.
For GPC~\citep{kurtz2025gpc}, we adapt the method to the same offline setting and use the same pretrained LeWM dynamics model as WATO.
All generative methods use the same goals and execution budget.

\paragraph{Model-based control.}
CEM~\citep{maes2026lewm, kim2026wombet} optimizes action trajectories online using the
same pretrained world model and terminal task cost $J$ as WATO.
It uses a population of $300$ trajectories, retains $30$ elites
(an elite fraction of $0.1$), and performs $30$ optimization rounds for
PushT and Reacher and $10$ rounds for the OGBench manipulation
environments.
The initial standard deviation is $1$, no warm start is used, and the
final Gaussian mean is returned as the action plan.
Gradient-based MPC uses $8$ independently initialized action
trajectories, $40$ gradient steps, and a learning rate of $0.05$.
Neither method updates the world model during evaluation.

We use the released LeWM CEM hyperparameters without tuning.
Hyperparameters for learned baselines are selected on validation tasks and then fixed for test evaluation.

\subsection{Hyperparameters}
\label{app:experimental_hyperparameters}

We use the same WATO configuration across all environments.
The planning horizon is $H=5$ latent transitions with $B=5$ primitive actions per block.
At deployment, WATO generates $K=8$ candidates using Euler integration with NFE $=8$.
We set $\lambda_A=10$, $\lambda_Z=0.1$, and $\lambda_{\mathrm{task}}=5\times10^{-3}$.
The trajectory optimizer uses at most $200$ L-BFGS-B iterations during WATO training and is not used at deployment.
Ablations change only the component specified by each experiment.

\subsection{Success Criteria}
\label{app:success_criteria}

We evaluate success using environment-specific semantic state criteria rather than the latent-space terminal cost used for planning.
The latent cost ranks candidate action trajectories, while success is computed from simulator state.
\Cref{tab:success_criteria} summarizes the criterion for each environment.

An episode is successful if its criterion is satisfied at least once during execution.
WATO and the other WAF-based controllers terminate immediately upon success.
The released Reacher CEM parity evaluator is an exception: it continues until the full episode budget is exhausted but keeps the first observed success for evaluation.

\begin{table}[!htbp]
\centering
\caption{
\textbf{Environment-specific semantic success criteria.}
These criteria use privileged simulator state and are distinct from the pretrained world model's latent-space planning cost.
}
\label{tab:success_criteria}
\scriptsize
\setlength{\tabcolsep}{5pt}
\begin{tabular}{lp{0.77\textwidth}}
\toprule
Environment & Success criterion \\
\midrule
\texttt{TwoRooms}
&
The Euclidean distance between the agent position and target position is
less than $16$ coordinate units.
\\

\texttt{Reacher}
&
The maximum absolute joint-position error over the two target joints is
less than $0.05$ radians.
\\

\texttt{PushT}
&
The positional error over the first four task-state coordinates is less
than $20$, and the wrapped block-orientation error is less than
$\pi/9$ radians.
\\

\texttt{cube-single}
&
The Euclidean distance between the cube position and target position is
at most $0.04$ meters.
\\

\texttt{cube-double}
&
Both cubes are within $0.04$ meters of their respective targets,
i.e.,
$\max_{i\in\{1,2\}}\left\lVert p_i-p_i^{g}\right\rVert_2 \leq 0.04$.
\\

\texttt{scene-play}
&
The cube-position error is at most $0.04$ meters, the absolute drawer
and window position errors are each at most $0.04$, and every discrete
button state matches its target state.
All conditions must hold simultaneously.
\\

\texttt{puzzle-3x3}
&
All nine binary button states match their target states,
i.e.,
$\sum_{i=1}^{9}\mathbf{1}\!\left[s_i \neq s_i^{g}\right] = 0$.
\\
\bottomrule
\end{tabular}
\end{table}


\section{Full Results}
\label{app:full_results}

This section reports the per-environment results underlying the aggregate comparisons in the main paper.
We provide the full WATO ablations, the absolute success rates used to measure optimization amortization, and final goal errors.

\subsection{Full WATO Ablations}
\label{app:full_wato_ablations}

\Cref{tab:app_wato_ablation} reports the complete training-objective and reference-design ablations.
All variants use the same joint WAF architecture, offline data, pretrained world model, and evaluation protocol.

The training-objective ablation isolates how trajectory optimization enters learning.
Joint WAF uses flow matching alone.
Direct rollout adds task supervision through the world model, direct rollout with consistency additionally aligns generated states with the rollout, and distillation uses optimized trajectories as detached targets.
WATO achieves the highest average success rate, improving from $79.9\%$ for Joint WAF to $91.7\%$.

The reference-design ablation isolates the two components of the world-action trajectory.
Using only the action reference achieves $85.7\%$, while using only the state reference achieves $85.0\%$.
Their combination reaches $91.7\%$, showing that neither reference alone recovers full WATO performance.

Removing $d_Z$ eliminates both its dynamics-aware regularization and the generated future reference.
To separate these effects, the rollout-derived variant replaces $\tilde Z$ with $\mathcal R_\theta(z_t,\tilde A)$ while retaining both reference penalties.
The backward pass includes the dependence of this reference on $\tilde A$.
This retains dynamics-aware regularization without using the generated future as a control reference, as discussed in Appendix~\ref{app:local_geometry}.

The average success rate of $87.9\%$ place this variant $2.2$ pp above the action-reference-only variant and $3.8$ points below WATO.
This indicate that dynamics-aware regularization explains part, but not all, of the state-reference benefit.
The remaining gap support using generated futures to guide optimization beyond the consequences already predicted for the proposed actions.
Conversely, performance comparable to WATO suggest that rollout-derived references capture most of this benefit.

\begin{table}[!htbp]
\centering
\caption{
\textbf{Full WATO ablation results.}
Results are success rates (\%, mean $\pm$ std over $4$ seeds).
}
\label{tab:app_wato_ablation}
\scriptsize
\setlength{\tabcolsep}{4pt}
\begin{tabular}{lcccccccc}
\toprule
& \multicolumn{4}{c}{Training objective}
& \multicolumn{3}{c}{Reference design}
& \\
\cmidrule(lr){2-5}
\cmidrule(lr){6-8}
Environment
& Joint WAF
& Direct
& \shortstack{Direct +\\Consistency}
& Distillation
& Action only
& State only
& \shortstack{Rollout-derived\\reference}
& \textcolor{pubblue}{\textbf{WATO}} \\
\midrule
\texttt{TwoRooms}
& $83$ \std{1}
& $85$ \std{1}
& $87$ \std{2}
& $90$ \std{0}
& $86$ \std{2}
& $88$ \std{3}
& $88$ \std{3}
& $\mathbf{93}$ \std{0} \\

\texttt{Reacher}
& $67$ \std{3}
& $77$ \std{5}
& $79$ \std{4}
& $81$ \std{4}
& $79$ \std{3}
& $76$ \std{4}
& $81$ \std{2}
& $\mathbf{85}$ \std{2} \\

\texttt{PushT}
& $84$ \std{4}
& $94$ \std{3}
& $95$ \std{3}
& $93$ \std{2}
& $95$ \std{3}
& $91$ \std{4}
& $96$ \std{4}
& $\mathbf{100}$ \std{0} \\

\texttt{cube-single}
& $88$ \std{1}
& $96$ \std{2}
& $97$ \std{2}
& $96$ \std{1}
& $96$ \std{2}
& $94$ \std{2}
& $97$ \std{1}
& $\mathbf{99}$ \std{1} \\

\texttt{cube-double}
& $89$ \std{2}
& $87$ \std{3}
& $88$ \std{3}
& $86$ \std{2}
& $84$ \std{3}
& $85$ \std{3}
& $87$ \std{5}
& $\mathbf{90}$ \std{3} \\

\texttt{scene-play}
& $78$ \std{3}
& $90$ \std{4}
& $91$ \std{3}
& $91$ \std{3}
& $87$ \std{4}
& $89$ \std{4}
& $90$ \std{3}
& $\mathbf{94}$ \std{1} \\

\texttt{puzzle-3x3}
& $70$ \std{3}
& $71$ \std{4}
& $75$ \std{2}
& $79$ \std{3}
& $73$ \std{4}
& $72$ \std{4}
& $76$ \std{2}
& $\mathbf{81}$ \std{2} \\
\midrule
Average
& $79.9$
& $85.7$
& $87.4$
& $88.0$
& $85.7$
& $85.0$
& $87.9$
& $\mathbf{91.7}$ \\
\bottomrule
\end{tabular}
\end{table}

\subsection{Training Horizon and Test Goal Distance}
\label{app:horizon_goal_distance}

The main experiments use a training horizon $L_{\mathrm{train}}=25$ and test goals $D=25$ primitive steps ahead.
We additionally vary
$L_{\mathrm{train}}\in\{10,25,50,100\}$
and
$D\in\{10,25,50,100\}$
on \texttt{PushT}, \texttt{cube-single}, and \texttt{scene-play}.
For each goal distance $D$, the execution budget is $2D$ steps.
All other settings are unchanged.

These horizons correspond to physical durations.
In \texttt{PushT}, one environment step is $0.1$\,s, so
$L_{\mathrm{train}}=\{10,25,50,100\}$
corresponds to $\{1,2.5,5,10\}$\,s.
In \texttt{cube-single} and \texttt{scene-play}, one step is $0.05$\,s, giving
$\{0.5,1.25,2.5,5\}$\,s.
The execution budget is twice the test-goal duration.

\Cref{tab:horizon_goal_distance} shows that WATO is not limited to short local control.
The best training horizon generally grows with the test goal distance, although exact matching is not always optimal.
For example, $L_{\mathrm{train}}=50$ achieves $82\%$ success on \texttt{PushT} at $D=50$, corresponding to a $5$\,s trajectory.
An average \texttt{PushT} episode spans roughly $125$ steps, or $12.5$\,s, so a $50$-step WATO horizon already covers about $40\%$ of a full episode.
A complete task therefore spans only about two to three such horizons, leaving less work for a separate high-level planner than a purely local-controller view would suggest.

Performance still decreases as the temporal scale grows.
At $D=100$, the best success rates are $48\%$, $58\%$, and $52\%$ on
\texttt{PushT}, \texttt{cube-single}, and \texttt{scene-play}, respectively.
Longer tasks with more objects, stages, or distant goals may therefore benefit from hierarchy or high-level planning.
Such a planner, however, can operate over multi-second WATO controllers rather than primitive local actions.
These results further support the perspective in Appendix~\ref{app:wato_controller}: WATO is a strong default controller over a broader temporal range than the main $25$-step setting alone suggests.

\begin{table}[!htbp]
\centering
\caption{
\textbf{Training horizon and test goal distance.}
WATO success rates (\%) for different training horizons $L_{\mathrm{train}}$ and test goal distances $D$, measured in primitive environment steps.
Bold indicates the best training horizon for each environment and test goal distance.
}
\label{tab:horizon_goal_distance}
\scriptsize
\setlength{\tabcolsep}{8pt}
\begin{tabular}{llcccc}
\toprule
Environment
& $L_{\mathrm{train}} \backslash D$
& $10$
& $25$
& $50$
& $100$ \\
\midrule

\multirow{4}{*}{\texttt{PushT}}
& $10$  & $\mathbf{100}$ & $80$ & $52$ & $16$ \\
& $25$  & $74$ & $\mathbf{100}$ & $70$ & $30$ \\
& $50$  & $30$ & $66$ & $\mathbf{82}$ & $\mathbf{48}$ \\
& $100$ & $8$  & $24$ & $50$ & $40$ \\
\midrule

\multirow{4}{*}{\texttt{cube-single}}
& $10$  & $\mathbf{100}$ & $68$ & $32$ & $10$ \\
& $25$  & $70$ & $\mathbf{94}$ & $40$ & $24$ \\
& $50$  & $24$ & $58$ & $\mathbf{92}$ & $44$ \\
& $100$ & $5$  & $18$ & $42$ & $\mathbf{58}$ \\
\midrule

\multirow{4}{*}{\texttt{scene-play}}
& $10$  & $\mathbf{97}$ & $64$ & $28$ & $6$ \\
& $25$  & $68$ & $\mathbf{99}$ & $50$ & $14$ \\
& $50$  & $20$ & $55$ & $\mathbf{76}$ & $32$ \\
& $100$ & $4$  & $14$ & $35$ & $\mathbf{52}$ \\
\bottomrule
\end{tabular}
\end{table}

\subsection{Full Optimization-Amortization Results}
\label{app:full_amortization}

The main paper reports the additional success obtained by applying trajectory optimization before and after WATO training.
\Cref{tab:app_wato_amortization} reports the corresponding absolute success rates.

Before WATO, trajectory optimization raises average success from $79.9\%$ to $90.3\%$.
After WATO, the same optimizer raises success only from $91.7\%$ to $93.0\%$.
Thus, most of the performance previously obtained through online optimization is already present in the WATO generator.

\begin{table}[!htbp]
\centering
\vspace{0.5em}
\caption{
\textbf{Absolute performance before and after trajectory optimization.}
The table expands the pp gains reported in \Cref{tab:wato_amortization}.
Results are success rates (\%) from $100$ tests.
}
\label{tab:app_wato_amortization}
\scriptsize
\setlength{\tabcolsep}{5.5pt}
\begin{tabular}{lcccc}
\toprule
& \multicolumn{2}{c}{Before WATO}
& \multicolumn{2}{c}{After WATO} \\
\cmidrule(lr){2-3}
\cmidrule(lr){4-5}
Environment
& Joint WAF
& + Traj. Opt.
& WATO
& + Traj. Opt. \\
\midrule
\texttt{TwoRooms}
& $83$ & $91$ & $93$ & $94$ \\

\texttt{Reacher}
& $67$ & $82$ & $85$ & $88$ \\

\texttt{PushT}
& $84$ & $98$ & $100$ & $100$ \\

\texttt{cube-single}
& $88$ & $97$ & $99$ & $100$ \\

\texttt{cube-double}
& $89$ & $92$ & $90$ & $91$ \\

\texttt{scene-play}
& $78$ & $92$ & $94$ & $95$ \\

\texttt{puzzle-3x3}
& $70$ & $80$ & $81$ & $83$ \\
\midrule
Average
& $79.9$
& $90.3$
& $91.7$
& $93.0$ \\
\bottomrule
\end{tabular}
\end{table}

The absolute results also separate optimization quality from amortization.
Online trajectory optimization is already strong when applied to the imitation-trained generator, reaching $90.3\%$ average success.
WATO reaches $91.7\%$ without this online optimization, while an additional optimization step provides only a small remaining gain.
This supports the interpretation that WATO transfers the useful effect of trajectory optimization into generation rather than relying on a weak optimizer.

\subsection{Final Goal Error}
\label{app:final_goal_error}

Success is a thresholded measure and does not distinguish failures that terminate near the goal from those that remain far away.
We therefore also report the final latent goal error
$\|E_\theta(o_T)-g\|_2$,
where $o_T$ is the final observation after the execution budget.
Lower values indicate trajectories that terminate closer to the target.

\Cref{tab:app_final_goal_error} reports the metric for the same methods and start-goal pairs as \Cref{tab:main_results}.
We report results separately for each environment because latent scales need not be comparable across independently trained world models.

\begin{table}[!htbp]
\centering
\caption{
\textbf{Final latent goal error.}
Lower is better.
Results use the same start-goal pairs and execution budget as the success-rate evaluation.
}
\label{tab:app_final_goal_error}
\scriptsize
\setlength{\tabcolsep}{7pt}

\begin{tabular}{lccccc}
\toprule
Environment
& GCBC
& GCIVL
& GCIQL
& QRL
& Flow Policy \\
\midrule
\texttt{TwoRooms}
& $12.7$ \std{4.3}
& $13.6$ \std{4.7}
& $12.9$ \std{4.9}
& $11.9$ \std{4.2}
& $13.1$ \std{4.4} \\

\texttt{Reacher}
& $124.3$ \std{24.8}
& $114.5$ \std{35.2}
& $125.0$ \std{15.1}
& $149.5$ \std{24.8}
& $162.8$ \std{35.7} \\

\texttt{PushT}
& $26.3$ \std{4.8}
& $33.0$ \std{6.2}
& $35.4$ \std{6.3}
& $28.6$ \std{5.2}
& $23.6$ \std{4.5} \\

\texttt{cube-single}
& $14.0$ \std{4.2}
& $23.0$ \std{6.4}
& $20.5$ \std{5.6}
& $14.7$ \std{4.9}
& $12.7$ \std{4.5} \\

\texttt{cube-double}
& $17.1$ \std{4.9}
& $34.1$ \std{8.7}
& $26.2$ \std{7.4}
& $18.0$ \std{5.1}
& $17.4$ \std{5.6} \\

\texttt{scene-play}
& $3.5$ \std{1.6}
& $4.5$ \std{2.2}
& $5.0$ \std{1.1}
& $3.6$ \std{1.0}
& $3.7$ \std{1.1} \\

\texttt{puzzle-3x3}
& $20.8$ \std{5.8}
& $21.5$ \std{7.1}
& $26.2$ \std{7.6}
& $23.0$ \std{6.0}
& $19.8$ \std{5.8} \\
\bottomrule
\end{tabular}

\vspace{0.8em}

\begin{tabular}{lccccc}
\toprule
Environment
& Joint WAF
& GPC
& CEM
& Grad. MPC
& \textbf{WATO} \\
\midrule
\texttt{TwoRooms}
& $12.3$ \std{4.0}
& $11.5$ \std{4.1}
& $65.5$ \std{14.3}
& $10.6$ \std{4.1}
& $10.4$ \std{3.5} \\

\texttt{Reacher}
& $116.0$ \std{5.1}
& $110.8$ \std{4.0}
& $285.3$ \std{17.7}
& $112.7$ \std{4.7}
& $92.9$ \std{4.2} \\

\texttt{PushT}
& $22.1$ \std{4.6}
& $29.5$ \std{3.5}
& $70.0$ \std{14.9}
& $18.8$ \std{4.0}
& $9.3$ \std{3.3} \\

\texttt{cube-single}
& $11.3$ \std{3.9}
& $13.4$ \std{4.6}
& $88.1$ \std{19.4}
& $13.8$ \std{4.5}
& $9.5$ \std{3.5} \\

\texttt{cube-double}
& $11.1$ \std{4.0}
& $16.8$ \std{6.1}
& $93.9$ \std{21.4}
& $15.0$ \std{4.5}
& $11.0$ \std{4.2} \\

\texttt{scene-play}
& $4.4$ \std{1.6}
& $4.2$ \std{2.8}
& $25.2$ \std{8.7}
& $5.3$ \std{2.9}
& $2.3$ \std{1.7} \\

\texttt{puzzle-3x3}
& $15.2$ \std{5.0}
& $25.4$ \std{7.2}
& $120.5$ \std{24.6}
& $17.7$ \std{5.2}
& $12.7$ \std{4.3} \\
\bottomrule
\end{tabular}
\end{table}

Final goal error complements success rate but does not replace it.
Success remains the primary metric because the evaluation criterion is defined in the environment, while latent distance measures how closely the final observation matches the goal under the shared representation.


\section{Deployment Efficiency Details}
\label{app:deployment_efficiency}

This section expands the deployment-efficiency comparison in Sec.~\ref{sec:wato_efficiency}.
We report per-environment latency, define the world model computation count, and study the effect of WATO sampling budget on success and deployment cost.

\subsection{Per-Environment Latency}
\label{app:latency}

We measure the wall-clock cost of one planning step.
WATO includes joint trajectory generation, world model evaluation of all generated action candidates, candidate selection, and action extraction.
CEM includes all online optimization iterations and world model rollouts.

\Cref{tab:app_latency} reports absolute deployment latency in seconds.
The corresponding relative speedups are reported in \Cref{tab:wato_efficiency}.
Absolute latency varies across environments, so we report per-environment values rather than averaging raw times.

\begin{table}[!htbp]
\centering
\caption{
\textbf{Per-environment deployment latency.}
Wall-clock time in seconds per planning step.
}
\label{tab:app_latency}
\scriptsize
\setlength{\tabcolsep}{7pt}
\begin{tabular}{lcc}
\toprule
Environment
& CEM (s)
& \textbf{WATO (s)} \\
\midrule
\texttt{TwoRooms}      & $1.909$ & $0.101$ \\
\texttt{Reacher}       & $1.999$ & $0.147$ \\
\texttt{PushT}         & $2.400$ & $0.096$ \\
\texttt{cube-single}   & $1.201$ & $0.207$ \\
\texttt{cube-double}   & $1.399$ & $0.222$ \\
\texttt{scene-play}    & $1.293$ & $0.106$ \\
\texttt{puzzle-3x3}    & $1.904$ & $0.207$ \\
\bottomrule
\end{tabular}
\end{table}

\subsection{World Model Evaluations}
\label{app:wm_evaluations}

Wall-clock latency depends on hardware and batching, so we also report world model computation directly.
We count one world model evaluation as one latent transition
$F_\theta(\mathcal C_h,A_h)$
for one candidate trajectory.
A full horizon-$H$ rollout therefore uses $H$ evaluations.

WATO generates $K$ action candidates and rolls each candidate through the world model once.
Its cost per planning step is therefore
\begin{equation}
    N_{\mathrm{WM}}^{\mathrm{WATO}}
    =
    KH.
    \label{eq:app_wato_wm_count}
\end{equation}

The flow matching-only joint WAF uses the same deployment procedure and therefore has the same world model count,
\begin{equation}
    N_{\mathrm{WM}}^{\mathrm{Joint\ WAF}}
    =
    KH.
    \label{eq:app_joint_waf_wm_count}
\end{equation}

Let $N_{\mathrm{CEM}}$ denote the number of CEM iterations and $M_{\mathrm{CEM}}$ the population size.
Each iteration evaluates all candidates over the full horizon, giving
\begin{equation}
    N_{\mathrm{WM}}^{\mathrm{CEM}}
    =
    N_{\mathrm{CEM}}
    M_{\mathrm{CEM}}
    H.
    \label{eq:app_cem_wm_count}
\end{equation}

For gradient-based MPC, let $N_{\mathrm{grad}}$ be the number of optimization steps and $M_{\mathrm{init}}$ the number of parallel initial action trajectories.
Each optimization step requires a forward rollout through the horizon, so
\begin{equation}
    N_{\mathrm{WM}}^{\mathrm{Grad.\ MPC}}
    =
    N_{\mathrm{grad}}
    M_{\mathrm{init}}
    H.
    \label{eq:app_grad_mpc_wm_count}
\end{equation}
Backward differentiation is included in wall-clock latency but not counted as an additional forward world model evaluation.

Because all compared methods use the same horizon $H$, the relative forward-model cost is independent of $H$.
For example, the CEM-to-WATO ratio is
\begin{equation}
    \frac{
        N_{\mathrm{WM}}^{\mathrm{CEM}}
    }{
        N_{\mathrm{WM}}^{\mathrm{WATO}}
    }
    =
    \frac{
        N_{\mathrm{CEM}}M_{\mathrm{CEM}}
    }{
        K
    }.
    \label{eq:app_cem_wato_ratio}
\end{equation}
This count is independent of batching: evaluating $K$ candidates in one batched forward pass still counts as $KH$ latent transitions.

\begin{table}[!htbp]
\centering
\caption{
\textbf{World model computation per planning step.}
Counts are latent transition evaluations.
}
\label{tab:app_wm_counts}
\scriptsize
\setlength{\tabcolsep}{5pt}
\begin{tabular}{lcc}
\toprule
Method
& Count
& Relative to WATO \\
\midrule
Joint WAF
& $KH$
& $1\times$ \\

CEM
& $N_{\mathrm{CEM}}M_{\mathrm{CEM}}H$
& $\dfrac{N_{\mathrm{CEM}}M_{\mathrm{CEM}}}{K}$ \\

Grad. MPC
& $N_{\mathrm{grad}}M_{\mathrm{init}}H$
& $\dfrac{N_{\mathrm{grad}}M_{\mathrm{init}}}{K}$ \\

\textbf{WATO}
& $KH$
& $\mathbf{1\times}$ \\
\bottomrule
\end{tabular}
\end{table}

\subsection{Success-Compute Tradeoff}
\label{app:success_compute}

WATO exposes two deployment-time compute parameters.
The candidate count $K$ controls how many generated action trajectories are evaluated by the world model, while NFE controls the cost of flow integration.
Increasing either can improve candidate quality or coverage at additional compute.

\begin{figure}[!htbp]
    \centering
    \includegraphics[width=0.75\linewidth]{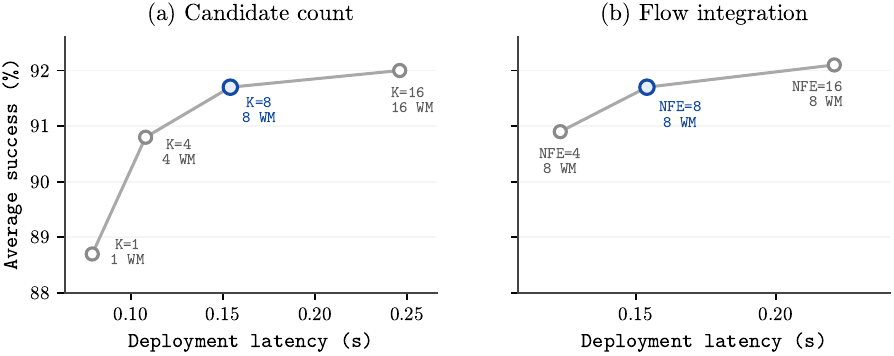}
    \caption{
    \textbf{WATO success-compute tradeoff.}
    \emph{Left:} varying the number of generated candidates $K$ with NFE $=8$.
    \emph{Right:} varying flow integration steps with $K=8$.
    The blue marker denotes the default setting used in the main experiments.
    }
    \label{fig:app_success_compute}
\end{figure}

We evaluate
$K\in\{1,4,8,16\}$
with NFE fixed at $8$, and
$\mathrm{NFE}\in\{4,8,16\}$
with $K$ fixed at $8$.
All other settings remain unchanged.
\Cref{fig:app_success_compute} reports average success across the $7$ environments against deployment latency, with the number of world model evaluations shown for each setting.

Increasing $K$ gives a clear gain from one to eight candidates, while the improvement from $K=8$ to $K=16$ is small despite the additional world model evaluations and latency.
Increasing NFE shows a similar pattern: NFE $=8$ captures most of the gain, with little improvement from doubling the integration steps to $16$.
We therefore use $K=8$ and NFE $=8$ as the default, which lies near the knee of both tradeoffs.


\section{Call for Research}
\label{app:call_for_research}
\titlespacing*{\paragraph}{0pt}{1.25ex plus .3ex minus .2ex}{1em}

WATO leaves further questions: \textcolor{pubblue}{\emph{what should we learn to do quickly, and what should we still plan online?}}
A strong local controller lets us ask where reasoning, value learning, and additional computation are actually most useful.

\paragraph{A strong controller is only the beginning.}
Long-horizon goal reaching has benefited from hierarchical learning~\citep{park2023hiql,park2025ogbench, kim2025roverfly}, but reliably composing local behaviors from offline visual data remains open.
Reaching waypoints is not enough if the resulting state makes the next transition difficult.
A high-level plan must choose reachable goals, leave the controller in useful states, and recover when execution deviates from the plan.
This is more than simply increasing the horizon.
We need work that local execution, subgoal selection, and their composition.

\paragraph{World action models beyond scale.}
WAM research need not be only about building larger foundation models.
As discussed in Appendix~\ref{app:wam_control_class}, joint WAMs also form a generative control model class whose objectives, references, and deployment rules deserve study.
How should generated futures influence actions?
When does joint generation help beyond an equally expressive action-only policy?
Which control principles transfer across architectures and scales?
Large-scale pretraining and control optimization address different questions.
Better generation and better control need not come from the same design choices.

\paragraph{Put values and broad knowledge where they help most.}
Our results also raise a question about where value learning belongs.
If local trajectories can already be generated reliably, values may be more useful for selecting subgoals and stitching behaviors than for extracting every motor action.
Hierarchical goal-conditioned learning provides a natural starting point~\citep{park2023hiql}.
High-level values should also reflect what the deployed low-level controller can actually execute, including its failures, rather than only what transitions appear in the dataset.

Large foundation models may similarly be most useful for semantic reasoning, task decomposition, and goal selection, while a dynamics-aware controller handles local execution.
SayCan illustrates how broad knowledge can be grounded through executable skills and their values~\citep{ahn2022saycan}.
At the same time, pretrained WAMs can directly generate low-level control~\citep{ye2026dreamzero}.
The question is therefore not only how much to scale, but \emph{where to put that scale}.

\paragraph{Act fast, deliberate when needed.}
\emph{Thinking fast and slow}~\citep{kahneman2011thinking} suggests allocating computation according to the decision.
A familiar subgoal may need only one forward pass, while difficult contact may justify extra sampling or optimization.
A useful extension of WATO would decide when direct generation is enough and when to refine a trajectory, increase sampling, or reconsider the goal.
Progress, uncertainty, or disagreement between generated and predicted futures could provide such signals, but they should be validated against execution outcomes.
The goal is to spend extra computation when it actually improves control.

\paragraph{Will more compute solve planning, or exploit model error?}
Faster hardware, GPU parallelism, and better solvers may make online trajectory optimization attractive.
Systems such as cuRobo already show large gains from parallel optimization~\citep{sundaralingam2023curobo}.
CEM is also not simply brute-force search: variants such as iCEM use temporal correlations and memory to improve sampling efficiency~\citep{pinneri2020icem}.
Future planners may further use learned proposals or references, erasing the distinction between online optimization and learned generation.

The harder question is whether computation or model error becomes the bottleneck.
More search can reduce predicted cost without making the world model more accurate, and may even find trajectories that exploit prediction errors~\citep{yu2020mopo, kim2026distribution}.
WATO's imitation loss and generated references guide optimization, but they do not guarantee model accuracy either.
A generator can also learn model exploitation.
Optimization error and model error therefore matter for both online planning and amortized control, as reflected in Appendix~\ref{app:optimization_amortization}.

\begin{center}
\colorbox{corebg}{
\parbox{0.95\linewidth}{
\vspace{0.35em}
\centering
\textcolor{pubblue}{
For local visual goals, \emph{WATO should be your favorite controller}.
\vspace{0.25em}
The next challenges are:
\\
\textbf{where to go}
$\;\bullet\;$
\textbf{when to think harder}
$\;\bullet\;$
\textbf{how to leverage local abilities in a long horizon}.
}
\vspace{0.35em}
}}
\end{center}


\end{document}